\documentclass{article}

\usepackage[preprint]{neurips_2026}

\usepackage[utf8]{inputenc}
\usepackage[T1]{fontenc}
\usepackage{hyperref}
\usepackage{url}
\usepackage{booktabs}
\usepackage{amsfonts}
\usepackage{amsmath}
\usepackage{nicefrac}
\usepackage{microtype}
\usepackage{graphicx}
\usepackage[table,xcdraw]{xcolor}  
\usepackage{multirow}
\usepackage{wrapfig}
\usepackage{subcaption}
\usepackage{tablefootnote}
\usepackage[most]{tcolorbox}
\usepackage{enumitem}
\definecolor{rowgray}{gray}{0.94}
\definecolor{overallgray}{gray}{0.87}
\tcbuselibrary{listings,breakable}

\title{One Token Away from Collapse: \\
The Fragility of Instruction-Tuned Helpfulness}

\author{Erfan Baghaei Potraghloo$^u$*, Seyedarmin Azizi$^u$*,
Souvik Kundu$^i$, and Massoud Pedram$^u$ \vspace{3mm}\\
$^u$University of Southern California,
Los Angeles, USA \\
$^i$Intel AI, USA\\
* Equal contribution authors\\
\texttt{\{baghaeip, seyedarm, pedram\}@usc.edu},  \texttt{souvikk.kundu@intel.com}\\
}

\begin{document}

\maketitle

\begin{abstract}
Instruction-tuned large language models produce helpful, structured responses, but how robust is this helpfulness under trivial constraints? We show that simple lexical constraints (banning a single punctuation character or common word) cause instruction-tuned LLMs to collapse their responses, losing 14--48\% of comprehensiveness across seven models spanning five families (7B--70B, open- and closed-weight). A blinded human evaluation with 10 STEM-trained evaluators confirms genuine content loss, with information criteria degrading $1.5$--$2.3\times$ more than surface criteria, a finding corroborated by over 4,100 automated pairwise comparisons (77--100\% baseline preference) across three LLM judges from two model families. Diagnostic analysis identifies this as a \emph{planning failure}: two-pass generation recovers 59--96\% of response length, and linear probes on prompt representations predict response length with $R^2 = 0.51$--$0.94$ before generation begins. The same probes yield negative $R^2$ on base models, confirming that instruction tuning introduces the representational structure underlying the collapse. Base models show no systematic degradation under identical constraints, demonstrating that instruction tuning couples task competence to narrow surface-form templates. The effect extends to realistic deployment constraints (preamble suppression, corporate tone guidelines, legal compliance hedging, accessibility requirements) causing comparable degradation ($-$22\% to $-$34\%), with suppressing the conversational opener alone (``Certainly!'') causing 40\% collapse on our most fragile model despite restricting only the opening tokens. We further show that standard independent LLM-as-judge evaluation detects only a 3.5\% quality drop where pairwise evaluation reveals 23\%, exposing a methodological blind spot in current evaluation practice.
\end{abstract}

\begin{figure*}[t]
\centering
\includegraphics[width=0.8\linewidth]{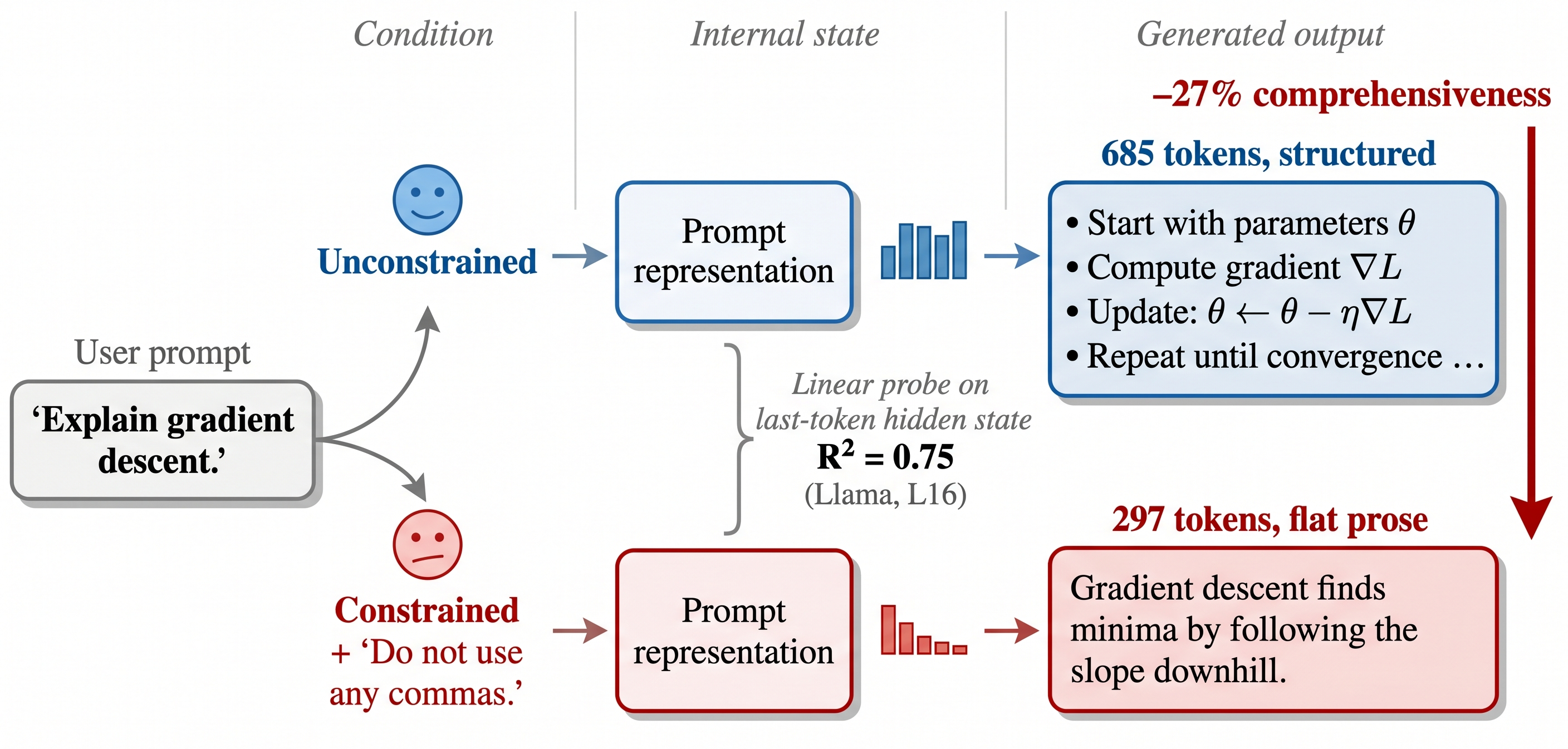}
\caption{\textbf{Constraint-induced response collapse.}
Adding a trivial lexical constraint (``do not use commas'') to an otherwise
identical prompt causes Llama-3.1-8B-Instruct to abandon its
structured 685-token response in favor of a 297-token flat-prose
summary, a 27\% loss in comprehensiveness despite no change in task or
knowledge requirements.}
\label{fig:schematic}
\end{figure*}

\section{Introduction}
\label{sec:intro}

Instruction tuning, the process of fine-tuning large language models (LLMs) on instruction--response pairs, often followed by preference optimization, is the standard recipe for producing helpful AI assistants~\citep{ouyang2022training, bai2022training, rafailov2024direct}. The implicit assumption is that instruction tuning teaches \emph{generalizable helpfulness}: the ability to provide thorough, accurate, and comprehensive responses regardless of surface-level formatting or stylistic choices. A well-aligned model should be able to explain gradient descent equally well whether or not it uses bullet points, commas, or any particular token. 

We challenge this assumption. We show that trivially constraining an instruction-tuned model's surface form (Figure~\ref{fig:schematic}: for instance, adding ``Do not use any commas in your response'' to a question) causes the model to \emph{collapse} its response. The model does not simply rephrase the same content without commas; it mode-switches to a drastically shorter, less helpful response. We quantify this through pairwise comprehensiveness ratings, how thoroughly a response covers the topic in terms of depth, detail, and examples, which directly capture the substantive quality that constitutes helpfulness in informational tasks. For Llama-3.1-8B-Instruct, banning commas reduces comprehensiveness by 27\% in pairwise evaluation. For Qwen-2.5-7B-Instruct, the same constraint reduces comprehensiveness by 61\%. The output remains free-form natural language. This is not a structured output problem. The model simply produces less. A blinded human evaluation (\S\ref{sec:human_eval}) with fine-grained criteria confirms the LLM-judge findings and establishes that information quality (coverage, comprehensiveness, helpfulness) degrades $1.5$--$2.3\times$ more than surface quality (conciseness, readability), ruling out length bias as an explanation. A length-invariant atomic-claim analysis (\S\ref{sec:coverage}) further confirms the collapse reflects real semantic loss: constrained responses preserve only 49.8\% of baseline factual claims.

This effect, which we term \textbf{constraint-induced response collapse}, is not a capability limitation. When we let the model generate freely and then rewrite its response under the constraint (a two-pass approach), it retains 59--96\% of the original length. The model \emph{can} write comprehensive comma-free prose; it just does not \emph{plan} to when the constraint is presented alongside the question. Linear probes on the model's prompt representations reveal that response length is predictable with $R^2 = 0.51$--$0.94$ from hidden states at middle layers, \emph{before a single token is generated}, with $R^2$ tracking collapse severity across five models from four families. The same probes applied to base (non-instruction-tuned) models yield negative $R^2$ at every layer, confirming that instruction tuning introduces both the behavioral collapse and the representational signature that accompanies it.

The critical evidence comes from comparing instruction-tuned models with their base (non-instruction-tuned) counterparts. Under identical constraints, base models show small, inconsistent effects (baseline win rates of 45--59\%, near or at chance), with some constraints actually \emph{improving} output quality. Instruction tuning transforms this noisy landscape into systematic, severe collapse (77--100\% baseline win rates). The fragility is not inherent to language models or to the constraints themselves; it is a specific artifact of instruction tuning, which couples task competence to narrow surface-form templates. Importantly, neither scale nor architecture resolves the fragility: Llama-3.3-70B-Instruct collapses \emph{more} than Llama-3.1-8B ($-$34.7\% vs.\ $-$25.9\%), the closed-weight GPT-4o-mini suffers 31\% loss, and the effect replicates across all seven instruction-tuned models we evaluate (\S\ref{sec:scaling}).

We additionally uncover a methodological blind spot: standard independent LLM-as-judge scoring~\citep{zheng2024judging}, the dominant evaluation paradigm, detects only a 3.5\% average quality drop from constraints that cause 23\% degradation in pairwise comparison on the same model with the same judge, a $6.7\times$ gap. This suggests that the evaluation community may be systematically underestimating quality loss in constrained generation settings.

\paragraph{Contributions.}
\begin{enumerate}[leftmargin=1.5em, itemsep=1pt]
    \item We document \textbf{constraint-induced response collapse}: 14--48\% comprehensiveness loss across seven instruction-tuned models spanning five families, 7B--70B scale, and open-/closed-weight systems, established via blinded human evaluation, 4,100+ pairwise comparisons, and a length-invariant atomic-claim analysis showing only 49.8\% of baseline claims preserved (\S\ref{sec:collapse}).
    \item We provide converging evidence that the collapse is a \textbf{planning failure}: two-pass recovery (59--96\%), a predictive representational signature ($R^2 = 0.51$--$0.94$), and base-model controls (negative $R^2$) (\S\ref{sec:mechanistic}).
    \item We show \textbf{instruction tuning specifically introduces} this fragility; base models show only small, inconsistent effects (\S\ref{sec:instruct}).
    \item We demonstrate that \textbf{independent LLM-as-judge evaluation is blind} to the collapse, detecting $<$20\% of pairwise-measured quality loss (\S\ref{sec:deployment_eval}).
    \item The effect \textbf{extends to realistic deployment constraints}: preamble suppression, corporate tone, legal hedging, and accessibility requirements cause 13--50\% degradation (\S\ref{sec:deployment_eval}).
\end{enumerate}

\noindent This paper is diagnostic: we establish and characterize the collapse, identify instruction tuning as its source, and show that models retain latent capability. Developing algorithmic mitigations remains future work.

\section{Related Work}
\label{sec:related}

\paragraph{Structured output and the format tax.}
\citet{tam2024speak} first documented that restricting LLMs to structured formats (JSON, XML, YAML) degrades reasoning accuracy by 10--15\%. Concurrent work by \citet{le2025formattax} systematically measures this ``format tax'' across ten models and four formats, decomposing degradation into prompt-level (dominant) and decoder-level (minor) components, and finding that closed-weight models largely resist the format tax. Our work differs in three key ways: (i)~our constraints are \emph{lexical} (token bans), not structural (format changes), and the output remains free-form natural language; (ii)~our effect sizes are substantially larger (14--48\% vs.\ 5--15\%); and (iii)~we find that closed-weight models are \emph{not} protected against lexical constraint collapse: GPT-4o-mini shows 31\% comprehensiveness loss. We additionally provide diagnostic analysis (planning failure, representational signature) and base-vs-instruct comparison that prior work does not. \citet{deco2025} propose decoupled generation to separate formatting from reasoning; their approach is complementary to our analysis.

\paragraph{Instruction following and constraint satisfaction.}
A growing body of work evaluates whether LLMs satisfy explicit constraints: IFEval~\citep{zhou2023instruction} introduced verifiable instruction-following benchmarks, and subsequent work has expanded to multi-constraint~\citep{he2024complex, wen2024benchmarking}, multi-turn~\citep{he2024multiif}, and agentic~\citep{agentif2025} settings. These benchmarks measure constraint \emph{satisfaction rates} but do not assess what happens to response \emph{quality} when constraints are followed. Our work is orthogonal; we show that even when models successfully satisfy a constraint (e.g., 99\% comma avoidance), the quality of the satisfying response collapses. Training methods that improve constraint satisfaction~\citep{dong2025autoif, iopo2025} address a complementary problem.

\paragraph{Constrained decoding.}
Grammar-constrained decoding (GCD)~\citep{willard2023efficient, geng2024grammar} masks invalid tokens to guarantee structural compliance, while grammar-aligned decoding~\citep{park2024grammar} corrects the distribution distortion GCD introduces. These operate at the \emph{decoder level} with formal grammars. Our constraints operate at the \emph{prompt level}: we do not mask tokens; we ask the model to avoid them via natural language instruction. \citet{le2025formattax} show that prompt-level effects dominate decoder-level effects for structured output; we extend this finding to show that prompt-level lexical constraints produce even larger degradation.

\paragraph{Activation steering for instruction following.}
\citet{stolfo2025improving} use representation engineering to improve instruction-following accuracy via steering vectors. Their work focuses on increasing constraint \emph{satisfaction rates} and does not study quality degradation. Our findings are complementary: understanding the representational basis of collapse could inform better steering interventions.

\paragraph{Robustness and prompt sensitivity.}
Prior work has studied LLM sensitivity to prompt phrasing~\citep{sclar2023quantifying, mizrahi2024state, ifeval2025plus}, showing that semantically equivalent reformulations can change model behavior. Our setting is distinct: constraints are not paraphrases of the same instruction but rather explicit additions that change the task. The model correctly interprets the constraint; it simply cannot maintain quality while following it.

\section{Experimental Setup}
\label{sec:setup}

\paragraph{Prompts.} We construct a diverse evaluation set of 40 prompts spanning four categories: explanation/education, how-to/advice, analysis/comparison, and technical/detailed, with 10 prompts per category (full list in Appendix~\ref{app:prompts}).

\paragraph{Constraints.} We define eight lexical constraints organized into three types: \emph{Punctuation-level:} ban commas, colons, or semicolons. \emph{Pattern-level:} ban bullet points, numbered lists, and dashes. \emph{Word-level:} ban the word ``the'' or ban discourse markers (``however,'' ``therefore,'' etc.). We additionally test two compositional constraints: commas+colons and commas+bullets. Each constraint is appended to the prompt as a natural-language instruction. The output format remains free-form natural language; no structured output (JSON, XML) is requested. Full constraint definitions appear in Appendix~\ref{app:constraints}.

\paragraph{Models.} We evaluate models spanning five families, multiple scales, and both open- and closed-weight systems. \emph{Primary open-weight instruction-tuned (detailed analysis):} Llama-3.1-8B-Instruct~\citep{grattafiori2024llama}, Qwen-2.5-7B-Instruct~\citep{yang2025qwen}, and Mistral-7B-Instruct-v0.3~\citep{jiang2023mistral}. \emph{Extended instruction-tuned (behavioral evaluation):} Llama-3.3-70B-Instruct~\citep{grattafiori2024llama}, Qwen3-30B-A3B-Instruct~\citep{qwen3_2025} (MoE, 30B total / 3B active), OLMo-3-7B-Instruct~\citep{olmo2025} (fully open), and GPT-4o-mini~\citep{openai2024gpt4o} (closed-weight). \emph{Open-weight base:} Llama-3.1-8B, Qwen-2.5-7B, and Mistral-7B. Open-weight models use bfloat16 precision.

\paragraph{Generation and evaluation.} For each prompt--constraint pair, we generate three independent samples (temperature 0.7, top-$p$ 0.9, max 1024 tokens). We employ three evaluation protocols: (i)~\emph{independent scoring} (judge rates each response in isolation on four dimensions, 1--10 scale), (ii)~\emph{pairwise comparison} (judge sees both baseline and constrained responses side-by-side with positions randomized, rates on comprehensiveness and usefulness), and (iii)~\emph{blinded human evaluation} (10 STEM-trained evaluators rate blinded pairs on six fine-grained criteria; \S\ref{sec:human_eval}). We use GPT-4o-mini, GPT-4o, and Claude Sonnet 4.6 as judges. We report \emph{comprehensiveness change (\%)} and \emph{baseline win rate}. Comprehensiveness and usefulness track near-perfectly ($r = 0.94$--$1.00$) across all 13 model--judge configurations. Constraint satisfaction rates are $>$90\% for most constraints (Appendix~\ref{app:satisfaction}), confirming that the collapse occurs among responses that \emph{successfully follow} the constraint. All judge prompts are provided in Appendix~\ref{app:judge_prompts}.

\section{Constraint-Induced Response Collapse}
\label{sec:collapse}

\subsection{Main Results: Pairwise Comparison}
\label{sec:main_results}

\begin{figure}[t]
\centering
\includegraphics[width=\linewidth]{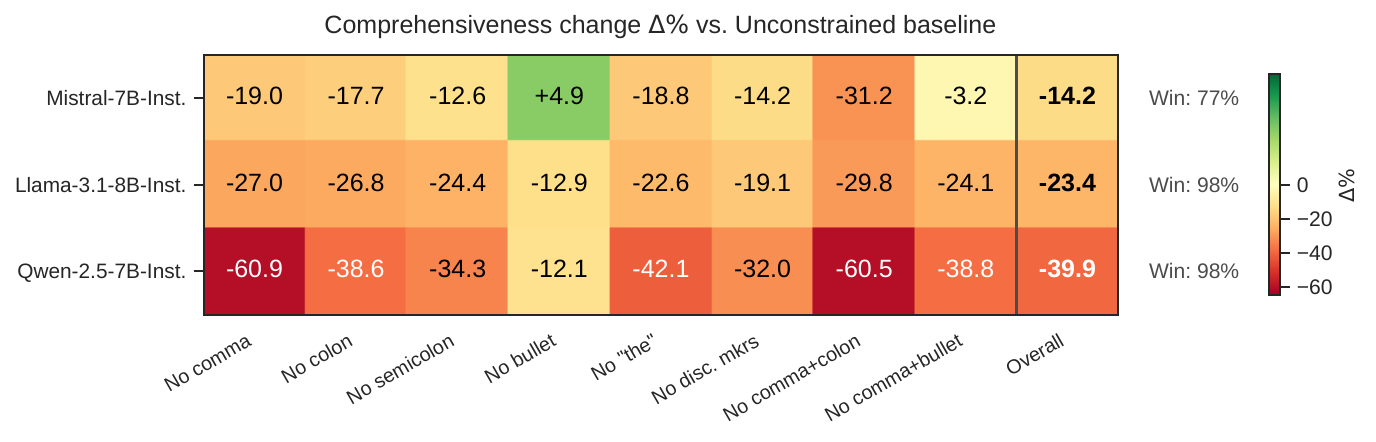}
\caption{\textbf{Pairwise comprehensiveness evaluation.} Heatmap of relative change $\Delta\%$ vs.\ unconstrained baseline (GPT-4o-mini judge); 40 prompts $\times$ 8 constraints = 320 pairs per model. Darker red indicates larger collapse. Baseline wins 97.5\% / 98.4\% / 77.2\% of pairs for Llama / Qwen / Mistral. Complete per-constraint numerical results with absolute scores appear in Appendix~\ref{app:full_results}.}
\label{fig:main_results}
\end{figure}

Figure~\ref{fig:main_results} presents the central finding: instruction-tuned models systematically lose comprehensiveness under lexical constraints. The effect is consistent across all three primary open-weight model families, all eight constraint types, and all four prompt categories (per-category breakdown in Appendix~\ref{app:category}).


Several patterns emerge. First, \textbf{the effect is large and consistent}: all three instruct models show substantial comprehensiveness loss (14--40\% overall), with the unconstrained baseline winning 77--98\% of pairwise comparisons. Second, \textbf{models differ in fragility}: Qwen is most fragile ($-$39.9\%), followed by Llama ($-$23.4\%) and Mistral ($-$14.2\%), suggesting that collapse severity varies with instruction-tuning recipe. Third, \textbf{the effect is not specific to formatting tokens}: banning the word ``the'' ($-$22.6\% on Llama, $-$42.1\% on Qwen, $-$18.8\% on Mistral) causes comparable damage to banning commas ($-$27.0\%, $-$60.9\%, $-$19.0\%), even though ``the'' plays no formatting role. This indicates the fragility is tied to token frequency and template disruption, not to formatting specifically. Compositional constraints show a collapse floor: banning commas and colons together ($-$29.8\% on Llama) produces only marginally worse results than banning commas alone ($-$27.0\%), suggesting a discrete strategy switch rather than continuous degradation.

\paragraph{Cross-judge and cross-family validation.}
To verify judge-independence, we repeat the pairwise evaluation with GPT-4o and Claude Sonnet 4.6 (a non-GPT-family model). All three judges detect the collapse with consistent severity ordering (Mistral $<$ Llama $<$ Qwen). Claude Sonnet 4.6, despite calibrating baseline scores lower (7.1--8.2 vs.\ 8.7--9.2 for GPT-4o-mini), detects degradation of $-$18.2\% to $-$48.9\%, closely matching or exceeding GPT-4o (full table in Appendix~\ref{app:cross_judge}). This rules out GPT-family judge biases. The collapse also replicates on MT-Bench (80 questions, 8 categories) with consistent per-constraint patterns (Appendix~\ref{app:mtbench}).

\subsection{Scale, Architecture, and Training Recipe Do Not Help}
\label{sec:scaling}

We evaluate four additional instruction-tuned models under identical constraints, all judged by GPT-4o: GPT-4o-mini~\citep{openai2024gpt4o} (closed-weight), Llama-3.3-70B-Instruct ($9\times$ larger), Qwen3-30B-A3B-Instruct (MoE, 30B total / 3B active), and OLMo-3-7B-Instruct (fully open).

\begin{table}[t]
\caption{\textbf{Constraint-induced collapse across seven instruction-tuned models} (GPT-4o pairwise judge). Every model collapses regardless of family, scale (7B to 70B), architecture (dense vs.\ MoE), or training openness. Larger models collapse \emph{more}, not less (Llama 8B $\to$ 70B: $-$25.9\% $\to$ $-$34.7\%).}
\label{tab:all_models}
\centering
\small
\begin{tabular}{l l r cc}
\toprule
Model & Family & Size & $\Delta$\% & Win\% \\
\midrule
Mistral-7B-Instruct-v0.3  & Mistral    &  7B         & $-$17.4 & 78.1 \\
Llama-3.1-8B-Instruct     & Llama      &  8B         & $-$25.9 & 97.5 \\
GPT-4o-mini                & GPT        & closed      & $-$31.0 & 99.1 \\
Llama-3.3-70B-Instruct    & Llama      & 70B         & $-$34.7 & 99.4 \\
OLMo-3-7B-Instruct        & OLMo       &  7B         & $-$37.7 & 95.9 \\
Qwen3-30B-A3B-Instruct    & Qwen~3 (MoE) & 30B (3B act.) & $-$37.9 & 99.1 \\
Qwen-2.5-7B-Instruct      & Qwen~2.5   &  7B         & $-$48.1 & 99.7 \\
\bottomrule
\end{tabular}
\end{table}

All seven instruction-tuned models collapse (Table~\ref{tab:all_models}). \textbf{Scale does not help:} Llama-3.3-70B collapses \emph{more} than Llama-3.1-8B ($-$34.7\% vs.\ $-$25.9\%). \textbf{Proprietary recipes are not immune:} GPT-4o-mini collapses at $-$31.0\% (99\% win rate), contrary to \citet{le2025formattax} who found closed-weight models resist the format tax on structured output. \textbf{Fully open training does not help:} OLMo collapses at $-$37.7\%. Per-constraint breakdowns for GPT-4o-mini appear in Appendix~\ref{app:gpt4omini}. All detailed numerical results are collected in Appendix~\ref{app:detailed_results}.

\subsection{The Collapse Reflects Semantic Loss, Not Verbosity Reduction}
\label{sec:coverage}

A natural concern is whether constrained responses are shorter but equally informative (``verbosity tax''). We conduct a length-invariant content analysis: for 8 stratified prompts across all three instruct models (192 pairs), GPT-4o extracts 11--20 atomic factual claims from each unconstrained response, then checks (with generous paraphrase matching) whether each claim survives in the constrained response. Coverage is length-invariant by construction.


\begin{wrapfigure}{r}{0.6\textwidth}
\vspace{-1.0em}
\centering
\includegraphics[width=\linewidth]{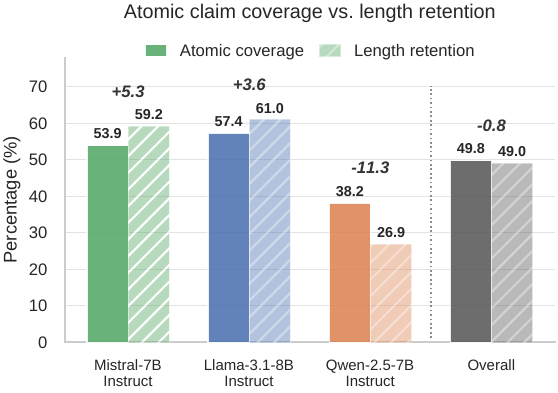}
\vspace{-0.8em}
\caption{\textbf{Atomic claim coverage analysis.} GPT-4o extracts factual claims from unconstrained responses and checks which survive in constrained responses. Coverage and length retention move together (gap $-0.8$pp), inconsistent with a pure verbosity account. 192 pairs, 3{,}355 atom checks. Numerical values in Appendix~\ref{app:coverage_table}.}
\label{fig:atomic_claims}
\vspace{-1.0em}
\end{wrapfigure}

Constrained responses preserve only 49.8\% of baseline factual claims on average (Figure~\ref{fig:atomic_claims}). Three findings are inconsistent with a pure verbosity account. First, coverage and length retention move together (overall gap $-0.8$pp): responses shed claims at approximately the same rate as words, whereas a verbosity-tax account would predict preserved coverage alongside reduced length. Second, Qwen exhibits a \emph{negative} gap ($-11.3$pp): its severely shortened responses are unusually dense per-word, yet still omit 62\% of baseline claims. Third, the \texttt{no\_bullet} constraint provides the cleanest anti-verbosity signal: averaged across models, responses retain 89\% of baseline length but only 58\% of claims (gap $+31$pp). Per-constraint detail appears in Appendix~\ref{app:coverage}; the coverage-analysis judge prompt is given in Appendix~\ref{app:coverage_prompts}.

\subsection{Human Evaluation}
\label{sec:human_eval}

To validate our LLM-based findings and disentangle genuine information loss from surface-level effects, we conduct a blinded human evaluation with 10 STEM-trained evaluators. Each evaluator receives blinded pairwise comparisons (positions randomized) and rates both responses on six dimensions (1--10): four \emph{information criteria} (semantic coverage, comprehensiveness, correctness, helpfulness) and two \emph{surface criteria} (verbosity, readability). We evaluate all 40 prompts under all 8 constraints for three models (960 pairwise comparisons, 5,760 individual ratings). Full protocol details and scoring rubrics appear in Appendix~\ref{app:human_eval}.

\begin{figure}[t]
\centering
\includegraphics[width=\linewidth]{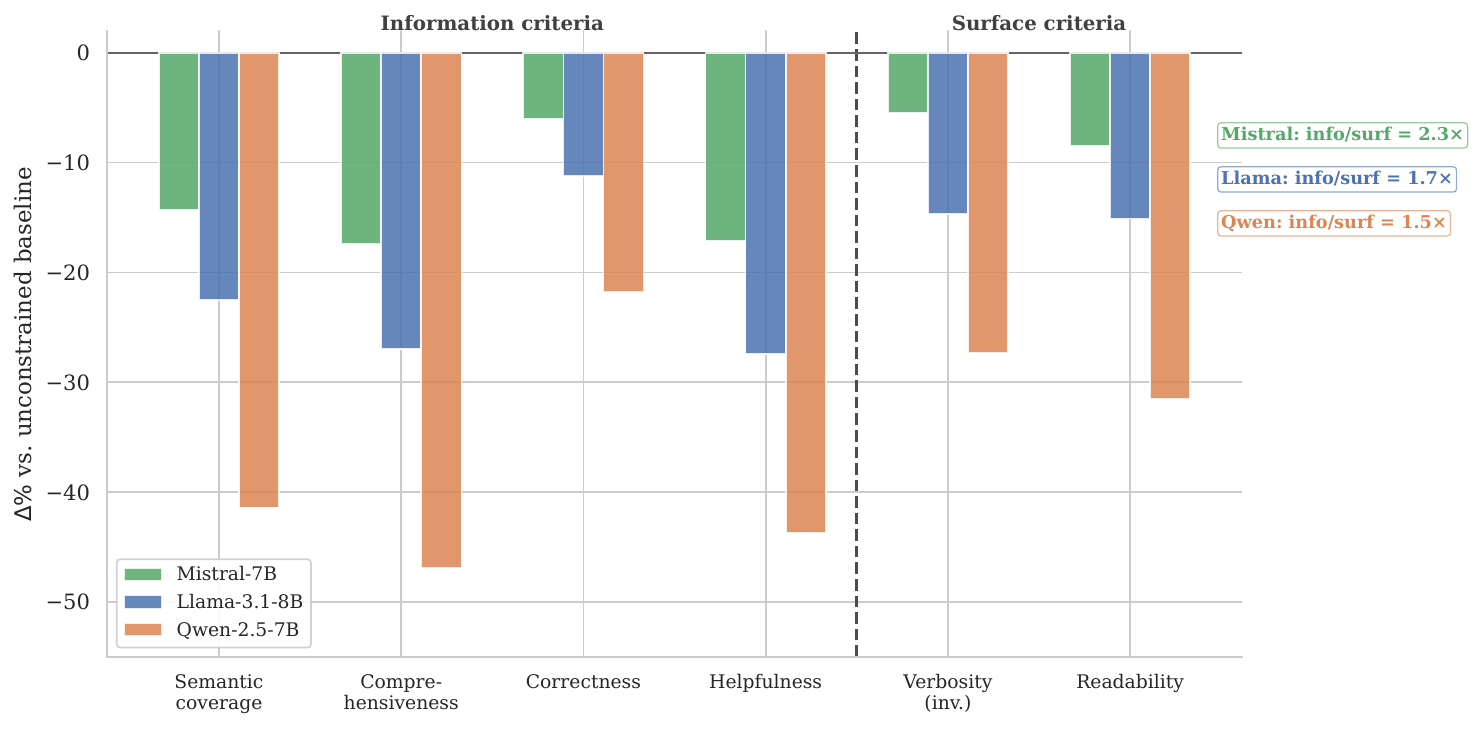}
\caption{\textbf{Human evaluation results.} Ten blinded evaluators rate responses on six criteria (1--10). Information criteria drop $1.5$--$2.3\times$ more than surface criteria, confirming genuine content loss. The dashed separator distinguishes information criteria (left) from surface criteria (right). 320 pairs per model. Numerical values in Appendix~\ref{app:human_eval_table}.}
\label{fig:human_eval}
\end{figure}

Human evaluators confirm and strengthen the LLM-judge findings (Figure~\ref{fig:human_eval}). Human-rated comprehensiveness drops ($-$17.4\% Mistral, $-$27.0\% Llama, $-$46.9\% Qwen) closely match LLM-judge ratings with identical severity ordering, and humans consistently detect \emph{larger} degradation. Information criteria drop $1.5$--$2.3\times$ more than surface criteria across all three models: Mistral provides the cleanest signal (16.3\% information loss vs.\ 7.0\% surface loss). If the collapse were driven by length preferences or formatting bias, surface and information drops would be comparable; they are not. Human-rated helpfulness and comprehensiveness change by nearly identical amounts ($-$27.4\% vs.\ $-$27.0\% on Llama, 0.4pp difference), validating comprehensiveness as the primary metric. Inter-rater standard deviation is 0.22--0.37, with effect sizes 3--15$\times$ the variability.

\section{Analysis: Why Does Collapse Happen?}
\label{sec:mechanistic}

The previous section establishes \emph{that} constraint-induced collapse occurs. We now investigate \emph{why}, through experiments replicated across three instruction-tuned models (Llama-3.1-8B, Qwen-2.5-7B, Mistral-7B).

\subsection{It Is a Planning Failure, Not a Capability Limitation}
\label{sec:twopass}

The collapse could reflect either (a)~the model genuinely cannot express complex content under constraints, or (b)~the model's planning mechanism selects a minimal response strategy despite having the capability to produce comprehensive constrained output. We distinguish these hypotheses using a two-pass protocol: for each of 10 prompts and 2 constraints (no comma, no ``the''), we generate a baseline response (no constraint), a single-pass response (constraint in prompt), and a two-pass response (generate baseline, then rewrite under the constraint with explicit instruction to maintain comprehensiveness).

\begin{table}[t]
\caption{\textbf{Two-pass recovery experiment.} Two-pass (generate freely, then rewrite under constraint) substantially recovers response length. The model \emph{can} produce comprehensive constrained output; it just does not plan to in single-pass. Retention measured relative to unconstrained baseline word count.}
\label{tab:twopass}
\centering
\small
\begin{tabular}{l ccc ccc}
\toprule
& \multicolumn{3}{c}{\textbf{Single-Pass Retention}} & \multicolumn{3}{c}{\textbf{Two-Pass Retention}} \\
\cmidrule(lr){2-4} \cmidrule(lr){5-7}
Constraint & Llama & Qwen & Mistral & Llama & Qwen & Mistral \\
\midrule
No comma     & 51\% &  9\% &  46\% & 100\% & 40\% &  86\% \\
No ``the''   & 46\% & 16\% &  57\% &  92\% & 79\% &  95\% \\
\midrule
\textbf{Overall} & \textbf{49\%} & \textbf{13\%} & \textbf{52\%} & \textbf{96\%} & \textbf{59\%} & \textbf{91\%} \\
\bottomrule
\end{tabular}
\end{table}

All three models confirm the planning failure hypothesis (Table~\ref{tab:twopass}). Llama and Mistral single-pass retains 49--52\% of baseline length, while two-pass retains 91--96\%, achieving near-perfect recovery. Qwen tells a more nuanced story: two-pass recovers substantially for ``no the'' (16\%$\to$79\%) but only partially for ``no comma'' (9\%$\to$40\%), indicating a secondary capability limitation for high-frequency syntactic tokens in more aggressively tuned models. A direct perplexity analysis (Appendix~\ref{app:perplexity}) rules out OOD likelihood failure as the primary driver: two-pass constrained text has only $1.15$--$1.51\times$ the base-model perplexity of unconstrained text for Llama and Mistral. Crucially, for Qwen, the collapsed single-pass response has \emph{higher} base-model perplexity ($5.6$) than the comprehensive two-pass rewrite ($5.1$), confirming the model is not defaulting to a higher-likelihood path. Qualitative two-pass examples are shown in Appendix~\ref{app:twopass_examples}.

\subsection{The Collapse Decision Is Encoded in Prompt Representations}
\label{sec:probing}

If the collapse is a planning failure, the decision should be detectable in the model's representations \emph{before generation begins}. For each of 40 prompts $\times$ 3 conditions (baseline, no comma, no ``the'') = 120 prompt variants, we extract hidden states at the last prompt token across five evenly-spaced layers and train Ridge regression probes via 5-fold cross-validation to predict response length. We focus on length prediction $R^2$ as the scientifically meaningful test, since constrained prompts contain additional text that makes classification trivial.

A simple linear probe explains 51--93\% of the variance in response length from the last prompt token's hidden state at the middle layer (${\sim}$50\% depth). All three models show a consistent layer profile: $R^2$ jumps sharply in early layers, peaks at ${\sim}$50\% depth, and plateaus or slightly decreases toward the final layer (per-layer details in Appendix~\ref{app:probing_layers}). This suggests that early-to-middle layers translate constraint detection into a response strategy, which is then maintained through the rest of the network. Token-level divergence analysis (Appendix~\ref{app:divergence}) confirms that models commit to a different response strategy within the first 1--3 tokens (JSD $= 0.46$--$0.54$).

\begin{wrapfigure}{r}{0.6\textwidth}
\vspace{-1.0em}
\centering
\includegraphics[width=\linewidth]{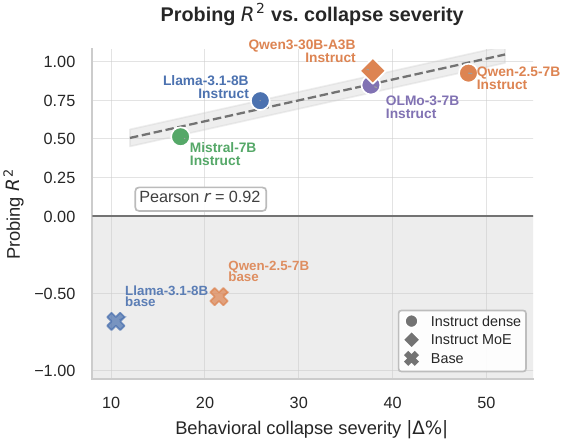}
\vspace{-0.8em}
\caption{\textbf{Probing $R^2$ tracks collapse severity.}
The collapse decision is encoded in prompt representations
\emph{before generation begins}: a linear probe on the last prompt token
predicts response length with $R^2 = 0.51$--$0.94$, with predictability
tracking collapse severity across five models ($r = 0.92$).
Base models yield negative $R^2$ (gray zone), confirming that instruction
tuning introduces both the behavioral collapse and its representational
signature. Numerical values in Appendix~\ref{app:probing_numerical}.}
\label{fig:probing_scatter}
\vspace{-3.0em}
\end{wrapfigure}

Probing $R^2$ correlates with behavioral collapse severity across all five instruction-tuned models (Figure~\ref{fig:probing_scatter}; numerical values in Appendix~\ref{app:probing_numerical}). Mistral, the most robust model ($-$17\%), yields $R^2 = 0.51$, while Qwen-2.5, the most fragile ($-$48\%), yields $R^2 = 0.93$, with a near-linear relationship ($r = 0.92$) across the five models. The convergence of three lines of evidence supports a causal interpretation: (i)~two-pass recovery demonstrates reversibility, (ii)~the same probes yield negative $R^2$ on base models sharing the same architecture, and (iii)~$R^2$ tracks collapse severity across five models from four families. Base models yield negative $R^2$ at every probed layer (Llama: $-$4.04; Qwen: $-$0.59), meaning the probe performs worse than predicting the mean. Instruction tuning simultaneously introduces the template-dependent response strategy and the predictive representational signature, both entirely absent in base models.



\section{Instruction Tuning Systematizes Fragility}
\label{sec:instruct}

We now test whether the fragility is inherent to language models or specific to instruction tuning by running the identical experiment on non-instruction-tuned (base) counterparts for all three families.



\begin{wrapfigure}{r}{0.6\textwidth}
\vspace{-1.0em}
\centering
\includegraphics[width=\linewidth]{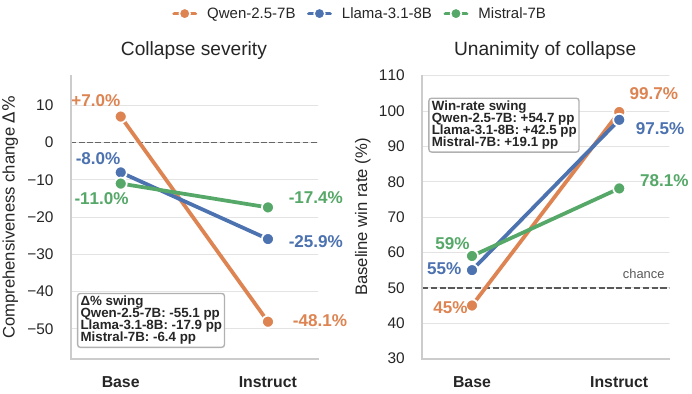}
\vspace{-0.8em}
\caption{\textbf{Instruction tuning creates the collapse.} Slope chart showing comprehensiveness change $\Delta\%$ (\textbf{left}) and baseline win rate (\textbf{right}) for base vs.\ instruction-tuned models under the same eight lexical constraints (GPT-4o pairwise judge; 320 pairs per model). Each line connects a base model to its instruction-tuned counterpart, making the within-family swing visually immediate: Qwen swings from $+7.0\%$ to $-48.1\%$ ($-55.1$ pp), Llama from $-8.0\%$ to $-25.9\%$ ($-17.9$ pp), and Mistral from $-11.0\%$ to $-17.4\%$ ($-6.4$ pp). Complete per-constraint numerical results in Appendix~\ref{app:full_results}.}
\label{fig:base_vs_instruct}
\vspace{-1.0em}
\end{wrapfigure}

Figure~\ref{fig:base_vs_instruct} reveals a striking contrast. Base models exhibit small, noisy effects: Qwen base shows $+$7.1\% \emph{improvement}, Llama base shows $-$7.5\% with a 55\% win rate (near chance), and Mistral base shows $-$11.2\% with a 59\% win rate. In contrast, all instruction-tuned models exhibit severe, systematic collapse ($-$17.4\% to $-$48.1\%). The same Qwen-2.5-7B architecture goes from constraints \emph{improving} output ($+$7.1\%) to catastrophic collapse ($-$48.1\%), a 55-percentage-point swing caused entirely by post-training alignment. The mechanism is template dependence: instruction tuning teaches a narrow repertoire of high-quality response templates. When constraints block key tokens these templates depend on, the model lacks an alternative comprehensive strategy and falls back to a minimal response mode.

\section{Deployment Constraints and Evaluation Blind Spots}
\label{sec:deployment_eval}

\subsection{Independent Evaluation Is Blind to Collapse}
\label{sec:eval}

Standard independent scoring dramatically underestimates constraint-induced quality loss. On Llama-3.1-8B-Instruct (GPT-4o-mini judge), independent scoring detects an average 3.5\% quality drop, while pairwise comparison reveals 23.4\%, a $6.7\times$ difference (per-constraint breakdown in Appendix~\ref{app:eval_blind}). For the no-bullet constraint, independent scoring detects \emph{zero} degradation despite 12.9\% pairwise-measured loss. Without seeing the full baseline response, the judge lacks a calibration reference and assigns inflated scores. Any evaluation of constrained generation systems that relies solely on independent scoring may systematically underestimate quality loss.

\subsection{The Constraint Tax in Deployment}
\label{sec:deployment}

The lexical constraints above are controlled diagnostic probes. We test whether analogous collapse occurs under production-grade constraints, evaluating Llama-3.1-8B-Instruct and Qwen-2.5-7B-Instruct under four enterprise-grade constraints (40 prompts, GPT-4o pairwise judge, 160 pairs per model): professional tone (brand guidelines), no preamble (API efficiency), hedging language (legal/compliance), and plain language (accessibility). Full constraint text appears in Appendix~\ref{app:deployment_constraints}.

All four deployment constraints cause significant comprehensiveness loss (Table~\ref{tab:deployment}), with the overall realistic-constraint collapse ($-$22.5\% Llama, $-$34.4\% Qwen) comparable to the lexical-constraint collapse ($-$25.9\%, $-$48.1\%). Three of the four constraints are fully unconfounded: they place no restriction on factual depth, vocabulary, response structure, or length. The no-preamble constraint is most striking: it restricts \emph{only} the opening tokens (``Certainly!'', ``Great question!''), yet Qwen loses 40.4\% of comprehensiveness and 74.7\% of word count (448$\to$113 words). This connects directly to the planning failure analysis: the RLHF-trained preamble functions as a conditional trigger that initializes the comprehensive response template. Extended discussion of each constraint type appears in Appendix~\ref{app:deployment_discussion}.

\begin{table}[t]
\caption{\textbf{Realistic deployment constraints cause collapse comparable to lexical bans.} Three constraints are fully unconfounded: professional tone, no-preamble, and hedging place zero restrictions on factual depth, vocabulary, or structure, yet cause 13--40\% degradation. The no-preamble constraint is most striking: suppressing only the conversational opener causes Qwen to lose 40\% comprehensiveness. $^\dagger$The plain language drop is partially confounded (see Appendix~\ref{app:deployment_discussion}).}
\label{tab:deployment}
\centering
\small
\begin{tabular}{l l cccc}
\toprule
& & \multicolumn{2}{c}{\textbf{Llama-8B}} & \multicolumn{2}{c}{\textbf{Qwen-7B}} \\
\cmidrule(lr){3-4} \cmidrule(lr){5-6}
Constraint & Real-world analogue & $\Delta$\% & Win\% & $\Delta$\% & Win\% \\
\midrule
Professional tone  & Brand/corporate tone     & $-$12.6 & 82 & $-$20.2 & 98 \\
No preamble        & API efficiency           & $-$17.6 & 92 & $-$40.4 & 100 \\
Hedging language   & Legal/compliance         & $-$26.4 & 95 & $-$26.8 & 100 \\
Plain language$^\dagger$ & Accessibility       & $-$33.1 & 100 & $-$50.1 & 100 \\
\midrule
\textbf{Overall (realistic)} & & $\mathbf{-22.5}$ & \textbf{92} & $\mathbf{-34.4}$ & \textbf{99} \\
\textit{Overall (lexical, ref.)} & & \textit{$-$25.9} & \textit{98} & \textit{$-$48.1} & \textit{100} \\
\bottomrule
\end{tabular}
\end{table}

\section{Discussion and Conclusion}
\label{sec:discussion}

Our results suggest that instruction tuning's apparent helpfulness is, at least in part, an artifact of learning a narrow distribution of response templates rather than developing generalizable competence. When any high-frequency token is banned, the model's learned templates become inaccessible and it defaults to a minimal response. The two-pass recovery result demonstrates that underlying knowledge and capability are intact; the planning mechanism simply cannot access them when the constraint is presented upfront. The probing results localize this failure to middle-layer representations (${\sim}$50\% depth), and the finding that $R^2$ tracks collapse severity suggests that reducing representational determinism at these layers may mitigate the collapse. The $6.7\times$ gap between independent and pairwise evaluation implies that deployed constrained generation systems may carry undetected quality loss. Combined with the deployment constraint results, these findings suggest the industry may be overestimating the utility of instruction-tuned models in constrained production environments. Additional discussion appears in Appendix~\ref{app:extended_discussion}.

\paragraph{Limitations.} Our diagnostic analysis is conducted on three 7--8B open-weight models; the behavioral signature is consistent across all seven instruct models (Table~\ref{tab:all_models}), but internal dynamics may differ in larger or closed-weight models. Our primary evaluation uses LLM judges validated by blinded human evaluation (\S\ref{sec:human_eval}). The probing analysis uses linear probes, which may underestimate representational complexity. Our constraint set, while diverse, is not exhaustive.

\paragraph{Conclusion.} Trivial lexical constraints cause 14--48\% comprehensiveness loss across seven instruction-tuned models, with blinded human evaluation confirming genuine content loss ($1.5$--$2.3\times$ more information than surface quality degradation). This is a planning failure, not a capability limitation: two-pass recovery reaches 59--96\%, the collapse is encoded in prompt representations ($R^2 = 0.51$--$0.94$), and base models show neither the behavioral nor representational signature. Realistic deployment constraints cause comparable degradation ($-$22\% to $-$34\%). We recommend pairwise evaluation for constrained generation and suggest that constraint robustness should become an explicit training objective. Promising directions include constraint-augmented alignment training~\citep{yuan2024length}, representation-level interventions~\citep{stolfo2025improving}, and training with diverse surface-form constraints to decouple competence from any particular template.


\bibliographystyle{neurips_2026}
\bibliography{neurips_2026}

\appendix

\section{Detailed Numerical Results}
\label{app:detailed_results}

This appendix collects the complete numerical results underlying the main-text figures and summary tables. Appendix~\ref{app:full_results} provides the full per-constraint pairwise evaluation tables for all models and judges. Appendix~\ref{app:probing_numerical} provides the numerical values behind Figure~\ref{fig:probing_scatter}. Appendix~\ref{app:cross_judge} presents cross-judge validation, Appendix~\ref{app:eval_blind} compares independent vs.\ pairwise evaluation sensitivity, Appendix~\ref{app:coverage_table} provides atomic claim coverage values behind Figure~\ref{fig:atomic_claims}, and Appendix~\ref{app:human_eval_table} provides human evaluation values behind Figure~\ref{fig:human_eval}.

\subsection{Full Per-Constraint Pairwise Results}
\label{app:full_results}

Tables~\ref{tab:app_llama_4omini}--\ref{tab:app_base} present the complete per-constraint pairwise comprehensiveness results for all models and judge configurations.

\begin{table}[h]
\caption{Per-constraint pairwise results: \textbf{Llama-3.1-8B-Instruct} (40 pairs per constraint).}
\label{tab:app_llama_4omini}
\centering
\small
\begin{tabular}{l cccc cccc}
\toprule
& \multicolumn{4}{c}{\textbf{GPT-4o-mini Judge}} & \multicolumn{4}{c}{\textbf{GPT-4o Judge}} \\
\cmidrule(lr){2-5} \cmidrule(lr){6-9}
Constraint & Base $\bar{c}$ & Const $\bar{c}$ & $\Delta$\% & Win\% & Base $\bar{c}$ & Const $\bar{c}$ & $\Delta$\% & Win\% \\
\midrule
\rowcolor{rowgray}
No comma       & 9.25 & 6.75 & $-$27.0 & 100 & 8.95 & 6.17 & $-$31.0 & 100 \\
No colon       & 9.22 & 6.75 & $-$26.8 & 100 & 8.93 & 6.40 & $-$28.3 & 100 \\
\rowcolor{rowgray}
No semicolon   & 9.22 & 6.97 & $-$24.4 &  98 & 8.90 & 6.62 & $-$25.6 &  98 \\
No bullet      & 9.07 & 7.90 & $-$12.9 &  85 & 8.78 & 7.30 & $-$16.8 &  88 \\
\rowcolor{rowgray}
No ``the''     & 9.28 & 7.17 & $-$22.6 & 100 & 8.88 & 6.72 & $-$24.2 &  98 \\
No disc.\ mkrs & 9.28 & 7.50 & $-$19.1 &  98 & 8.90 & 6.90 & $-$22.5 &  98 \\
\rowcolor{rowgray}
No comma+colon & 9.22 & 6.47 & $-$29.8 & 100 & 8.90 & 6.08 & $-$31.7 & 100 \\
No comma+bullet& 9.25 & 7.03 & $-$24.1 & 100 & 8.93 & 6.55 & $-$26.6 & 100 \\
\midrule
\rowcolor{overallgray}
\textbf{Overall} & \textbf{9.22} & \textbf{7.07} & $\mathbf{-23.4}$ & \textbf{97.5} & \textbf{8.89} & \textbf{6.59} & $\mathbf{-25.9}$ & \textbf{97.5} \\
\bottomrule
\end{tabular}
\end{table}

\begin{table}[h]
\caption{Per-constraint pairwise results: \textbf{Qwen-2.5-7B-Instruct} (40 pairs per constraint).}
\label{tab:app_qwen_4omini}
\centering
\small
\begin{tabular}{l cccc cccc}
\toprule
& \multicolumn{4}{c}{\textbf{GPT-4o-mini Judge}} & \multicolumn{4}{c}{\textbf{GPT-4o Judge}} \\
\cmidrule(lr){2-5} \cmidrule(lr){6-9}
Constraint & Base $\bar{c}$ & Const $\bar{c}$ & $\Delta$\% & Win\% & Base $\bar{c}$ & Const $\bar{c}$ & $\Delta$\% & Win\% \\
\midrule
\rowcolor{rowgray}
No comma       & 9.07 & 3.55 & $-$60.9 & 100 & 8.97 & 2.60 & $-$71.0 & 100 \\
No colon       & 9.20 & 5.65 & $-$38.6 & 100 & 8.97 & 5.00 & $-$44.3 & 100 \\
\rowcolor{rowgray}
No semicolon   & 9.18 & 6.03 & $-$34.3 & 100 & 8.97 & 5.25 & $-$41.5 & 100 \\
No bullet      & 9.07 & 7.97 & $-$12.1 &  88 & 8.95 & 7.30 & $-$18.4 &  98 \\
\rowcolor{rowgray}
No ``the''     & 9.20 & 5.33 & $-$42.1 & 100 & 8.97 & 4.42 & $-$50.7 & 100 \\
No disc.\ mkrs & 9.15 & 6.22 & $-$32.0 & 100 & 8.97 & 5.70 & $-$36.5 & 100 \\
\rowcolor{rowgray}
No comma+colon & 9.05 & 3.58 & $-$60.5 & 100 & 9.00 & 2.60 & $-$71.1 & 100 \\
No comma+bullet& 9.03 & 5.53 & $-$38.8 & 100 & 8.97 & 4.40 & $-$51.0 & 100 \\
\midrule
\rowcolor{overallgray}
\textbf{Overall} & \textbf{9.12} & \textbf{5.48} & $\mathbf{-39.9}$ & \textbf{98.4} & \textbf{8.97} & \textbf{4.66} & $\mathbf{-48.1}$ & \textbf{99.7} \\
\bottomrule
\end{tabular}
\end{table}

\begin{table}[h]
\caption{Per-constraint pairwise results: \textbf{Mistral-7B-Instruct-v0.3} (40 pairs per constraint).}
\label{tab:app_mistral}
\centering
\small
\begin{tabular}{l cccc cccc}
\toprule
& \multicolumn{4}{c}{\textbf{GPT-4o-mini Judge}} & \multicolumn{4}{c}{\textbf{GPT-4o Judge}} \\
\cmidrule(lr){2-5} \cmidrule(lr){6-9}
Constraint & Base $\bar{c}$ & Const $\bar{c}$ & $\Delta$\% & Win\% & Base $\bar{c}$ & Const $\bar{c}$ & $\Delta$\% & Win\% \\
\midrule
\rowcolor{rowgray}
No comma       & 8.70 & 7.05 & $-$19.0 &  83 & 8.40 & 6.42 & $-$23.5 & 85 \\
No colon       & 8.75 & 7.20 & $-$17.7 &  88 & 8.22 & 6.83 & $-$17.0 & 80 \\
\rowcolor{rowgray}
No semicolon   & 8.72 & 7.62 & $-$12.6 &  83 & 8.20 & 7.15 & $-$12.8 & 75 \\
No bullet      & 8.15 & 8.55 & $+$4.9  &  38 & 8.05 & 7.83 & $-$2.8  & 55 \\
\rowcolor{rowgray}
No ``the''     & 8.78 & 7.12 & $-$18.8 &  88 & 8.35 & 6.65 & $-$20.4 & 88 \\
No disc.\ mkrs & 8.80 & 7.55 & $-$14.2 &  85 & 8.30 & 6.97 & $-$16.0 & 80 \\
\rowcolor{rowgray}
No comma+colon & 8.90 & 6.12 & $-$31.2 &  98 & 8.47 & 5.50 & $-$35.1 & 90 \\
No comma+bullet& 8.53 & 8.25 & $-$3.2  &  58 & 8.28 & 7.38 & $-$10.9 & 73 \\
\midrule
\rowcolor{overallgray}
\textbf{Overall} & \textbf{8.67} & \textbf{7.43} & $\mathbf{-14.2}$ & \textbf{77.2} & \textbf{8.28} & \textbf{6.84} & $\mathbf{-17.4}$ & \textbf{78.1} \\
\bottomrule
\end{tabular}
\end{table}

\begin{table}[h]
\caption{Per-constraint pairwise results: \textbf{Base models} (GPT-4o judge, 40 pairs per constraint). \textbf{Bold} $\Delta\%$ and Win\% values indicate that the constrained response was rated higher than the unconstrained baseline, the opposite of what happens in instruction-tuned models.}
\label{tab:app_base}
\centering
\small
\begin{tabular}{l cccc cccc cccc}
\toprule
& \multicolumn{4}{c}{\textbf{Llama-3.1-8B (Base)}} & \multicolumn{4}{c}{\textbf{Qwen-2.5-7B (Base)}} & \multicolumn{4}{c}{\textbf{Mistral-7B (Base)}} \\
\cmidrule(lr){2-5} \cmidrule(lr){6-9} \cmidrule(lr){10-13}
Constraint & B & C & $\Delta$\% & W\% & B & C & $\Delta$\% & W\% & B & C & $\Delta$\% & W\% \\
\midrule
\rowcolor{rowgray}
No comma       & 6.50 & 5.33 & $-$18          & 62          & 5.83 & 5.72 & $-$2           & 55          & 5.97 & 5.47 & $-$8           & 58 \\
No colon       & 6.40 & 5.85 & $-$9           & 58          & 5.90 & 6.15 & $\mathbf{+4}$  & \textbf{48} & 5.92 & 5.30 & $-$11          & 58 \\
\rowcolor{rowgray}
No semicolon   & 6.55 & 5.22 & $-$20          & 75          & 5.88 & 6.28 & $\mathbf{+7}$  & \textbf{42} & 6.00 & 5.38 & $-$10          & 60 \\
No bullet      & 6.17 & 6.90 & $\mathbf{+12}$ & \textbf{38} & 5.70 & 6.62 & $\mathbf{+16}$ & \textbf{35} & 5.88 & 5.55 & $-$6           & 50 \\
\rowcolor{rowgray}
No ``the''     & 6.53 & 5.05 & $-$23          & 65          & 5.83 & 6.10 & $\mathbf{+5}$  & \textbf{48} & 6.20 & 4.72 & $-$24          & 70 \\
No disc.\ mkrs & 6.22 & 6.05 & $-$3           & 50          & 6.00 & 5.65 & $-$6           & 60          & 5.92 & 5.97 & $\mathbf{+1}$  & \textbf{50} \\
\rowcolor{rowgray}
No cm+colon    & 6.45 & 5.55 & $-$14          & 62          & 5.88 & 6.22 & $\mathbf{+6}$  & \textbf{48} & 6.05 & 5.12 & $-$15          & 62 \\
No cm+bullet   & 6.05 & 7.10 & $\mathbf{+17}$ & \textbf{30} & 5.53 & 7.08 & $\mathbf{+28}$ & \textbf{25} & 6.10 & 5.12 & $-$16          & 62 \\
\midrule
\rowcolor{overallgray}
\textit{Overall} & \textit{6.36} & \textit{5.88} & \textit{$-$8} & \textit{55} & \textit{5.82} & \textit{6.23} & \textbf{\textit{+7}} & \textbf{\textit{45}} & \textit{6.01} & \textit{5.33} & \textit{$-$11} & \textit{59} \\
\bottomrule
\end{tabular}
\end{table}

\subsection{Probing $R^2$ vs.\ Collapse Severity: Numerical Values}
\label{app:probing_numerical}

Table~\ref{tab:probing_numerical} provides the numerical values underlying Figure~\ref{fig:probing_scatter} in the main text.

\begin{table}[h]
\caption{\textbf{Probing $R^2$ tracks collapse severity, and base models show no representational signature.} The more severe the behavioral collapse, the more predictable the response length from prompt representations. The near-monotonic relationship across five models from four families is consistent with a shared planning-failure phenomenon. Base models yield negative $R^2$ at every layer, indicating response length is entirely unpredictable from representations. Instruction tuning introduces both the collapse and the representational signature.}
\label{tab:probing_numerical}
\centering
\small
\begin{tabular}{l l cc}
\toprule
Model & Family & Collapse ($\Delta$\%) & Probing $R^2$ (best) \\
\midrule
Mistral-7B-Instruct   & Mistral & $-$17.4\% & 0.514 \\
Llama-3.1-8B-Instruct & Llama   & $-$25.9\% & 0.747 \\
OLMo-3-7B-Instruct    & OLMo    & $-$37.7\% & 0.848 \\
Qwen3-30B-A3B-Instruct & Qwen~3  & $-$37.9\% & 0.940 \\
Qwen-2.5-7B-Instruct  & Qwen~2.5 & $-$48.1\% & 0.925 \\
\midrule
Llama-3.1-8B (base)    & Llama   & n/a & $-$4.04 \\
Qwen-2.5-7B (base)     & Qwen    & n/a & $-$0.59 \\
\bottomrule
\end{tabular}
\end{table}

\subsection{Cross-Judge and Cross-Family Validation}
\label{app:cross_judge}

\begin{table}[h]
\caption{\textbf{Cross-judge and cross-family validation.} Three judges from two model families (GPT-4o-mini, GPT-4o, Claude Sonnet 4.6) consistently detect the collapse with identical severity ordering. Claude Sonnet 4.6, despite assigning lower baseline scores (7.1--8.2 vs.\ 8.7--9.2 for GPT-4o-mini), detects comparable or larger degradation, ruling out GPT-family judge bias.}
\label{tab:cross_judge}
\centering
\small
\begin{tabular}{l cc cc cc}
\toprule
& \multicolumn{2}{c}{\textbf{GPT-4o-mini}} & \multicolumn{2}{c}{\textbf{GPT-4o}} & \multicolumn{2}{c}{\textbf{Claude Sonnet 4.6}} \\
\cmidrule(lr){2-3} \cmidrule(lr){4-5} \cmidrule(lr){6-7}
Model & $\Delta$\% & Win\% & $\Delta$\% & Win\% & $\Delta$\% & Win\% \\
\midrule
Mistral-7B-Instruct   & $-$14.2 & 77.2 & $-$17.4 & 78.1 & $-$18.2 & 71.9 \\
Llama-3.1-8B-Instruct & $-$23.4 & 97.5 & $-$25.9 & 97.5 & $-$28.4 & 96.9 \\
Qwen-2.5-7B-Instruct  & $-$39.9 & 98.4 & $-$48.1 & 99.7 & $-$48.9 & 99.7 \\
\bottomrule
\end{tabular}
\end{table}

\subsection{Independent vs.\ Pairwise Evaluation}
\label{app:eval_blind}

\begin{table}[h]
\caption{\textbf{Independent vs.\ pairwise evaluation} on Llama-3.1-8B-Instruct (GPT-4o-mini judge). Independent scoring detects $<\nicefrac{1}{5}$ of the quality loss measured by pairwise comparison.}
\label{tab:eval_blind}
\centering
\small
\begin{tabular}{l cc}
\toprule
Constraint & Independent Judge $\Delta$\% & Pairwise Judge $\Delta$\% \\
\midrule
No comma       & $-$5.4  & $-$27.0 \\
No colon       & $-$4.5  & $-$26.8 \\
No semicolon   & $-$3.2  & $-$24.4 \\
No bullet/lists& $-$0.0  & $-$12.9 \\
No ``the''     & $-$2.8  & $-$22.6 \\
No disc.\ markers & $-$0.5 & $-$19.1 \\
No comma+colon & $-$7.0  & $-$29.8 \\
No comma+bullet& $-$4.6  & $-$24.1 \\
\midrule
\textbf{Average} & $\mathbf{-3.5}$ & $\mathbf{-23.4}$ \\
\bottomrule
\end{tabular}
\end{table}

In independent evaluation, the judge assesses each response in isolation against an implicit quality standard. Constrained responses, while shorter and less comprehensive, are often \emph{locally coherent}: each sentence is accurate and well-formed. Without seeing the full baseline response, the judge lacks a calibration reference and assigns inflated scores. The baseline composite score (8.54/10) and the constrained composite (7.94--8.54/10) both fall in the ``good to very good'' range, masking the large gap in actual comprehensiveness.

\subsection{Atomic Claim Coverage: Numerical Values}
\label{app:coverage_table}

Table~\ref{tab:coverage_numerical} provides the numerical values underlying Figure~\ref{fig:atomic_claims} in the main text.

\begin{table}[h]
\caption{\textbf{Atomic claim coverage analysis} (numerical values for Figure~\ref{fig:atomic_claims}). GPT-4o extracts factual claims from unconstrained responses and checks which survive in constrained responses. Coverage and length retention move together (gap $-0.8$pp), inconsistent with a pure verbosity account. 192 pairs, 3{,}355 atom checks.}
\label{tab:coverage_numerical}
\centering
\small
\begin{tabular}{l ccc}
\toprule
Model & Atomic Coverage & Length Retention & Gap \\
\midrule
Mistral-7B-Instruct    & 53.9\% & 59.2\% & $+$5.3 pp \\
Llama-3.1-8B-Instruct  & 57.4\% & 61.0\% & $+$3.6 pp \\
Qwen-2.5-7B-Instruct   & 38.2\% & 26.9\% & $-$11.3 pp \\
\midrule
\textbf{Overall}       & \textbf{49.8\%} & \textbf{49.0\%} & $\mathbf{-0.8}$ \textbf{pp} \\
\bottomrule
\end{tabular}
\end{table}

\subsection{Human Evaluation: Numerical Values}
\label{app:human_eval_table}

Table~\ref{tab:human_eval_numerical} provides the numerical values underlying Figure~\ref{fig:human_eval} in the main text.

\begin{table}[h]
\caption{\textbf{Human evaluation results} (numerical values for Figure~\ref{fig:human_eval}). Ten blinded evaluators rate responses on six criteria (1--10). Information criteria drop $1.5$--$2.3\times$ more than surface criteria, confirming genuine content loss. 320 pairs per model.}
\label{tab:human_eval_numerical}
\centering
\small
\begin{tabular}{l l ccc}
\toprule
& & \textbf{Llama-8B} & \textbf{Mistral-7B} & \textbf{Qwen-7B} \\
\cmidrule(lr){3-3} \cmidrule(lr){4-4} \cmidrule(lr){5-5}
& Criterion & $\Delta$\% & $\Delta$\% & $\Delta$\% \\
\midrule
\multirow{4}{*}{\rotatebox{90}{\scriptsize Info}}
& Semantic coverage     & $-$22.5 & $-$14.3 & $-$41.4 \\
& Comprehensiveness     & $-$27.0 & $-$17.4 & $-$46.9 \\
& Correctness           & $-$11.2 & $-$6.0  & $-$21.8 \\
& Helpfulness           & $-$27.4 & $-$17.1 & $-$43.7 \\
\midrule
\multirow{2}{*}{\rotatebox{90}{\scriptsize Surf}}
& Verbosity (inv.)      & $-$14.7 & $-$5.5  & $-$27.3 \\
& Readability           & $-$15.1 & $-$8.5  & $-$31.5 \\
\midrule
\multicolumn{2}{l}{\textbf{Info criteria avg}} & $\mathbf{-25.6}$ & $\mathbf{-16.3}$ & $\mathbf{-44.0}$ \\
\multicolumn{2}{l}{\textbf{Surface criteria avg}} & $\mathbf{-14.9}$ & $\mathbf{-7.0}$ & $\mathbf{-29.4}$ \\
\multicolumn{2}{l}{\textbf{Info / Surface ratio}} & $\mathbf{1.7\times}$ & $\mathbf{2.3\times}$ & $\mathbf{1.5\times}$ \\
\multicolumn{2}{l}{Baseline win\% (coverage)} & 97\% & 78\% & 99\% \\
\bottomrule
\end{tabular}
\end{table}

\section{Evaluation Prompt List}
\label{app:prompts}

Our evaluation set consists of 40 prompts across four categories (10 per category), designed to elicit substantive, multi-paragraph responses that benefit from structured formatting.

\begin{tcolorbox}[colback=blue!3, colframe=blue!40!black, title={\textbf{Category 1: Explanation / Education} (10 prompts)}, breakable, fonttitle=\small]
\small
\begin{enumerate}[leftmargin=*, itemsep=2pt]
    \item Explain gradient descent in simple terms.
    \item What is photosynthesis and why is it important for life on Earth?
    \item How does a computer CPU process instructions?
    \item Explain the concept of supply and demand in economics.
    \item What is machine learning and how does it differ from traditional programming?
    \item Explain how vaccines work to protect against diseases.
    \item What is quantum computing and why is it potentially revolutionary?
    \item Explain the water cycle and its importance for the environment.
    \item How does encryption work to keep data secure?
    \item What is the theory of evolution and what evidence supports it?
\end{enumerate}
\end{tcolorbox}

\begin{tcolorbox}[colback=green!3, colframe=green!40!black, title={\textbf{Category 2: How-To / Advice} (10 prompts)}, breakable, fonttitle=\small]
\small
\begin{enumerate}[leftmargin=*, itemsep=2pt]
    \item How should I prepare for a technical job interview?
    \item What are the best practices for writing clean, maintainable code?
    \item How can I improve my public speaking skills?
    \item What steps should I take to start investing in the stock market?
    \item How do I write an effective research paper?
    \item What are good strategies for managing stress and anxiety?
    \item How should I approach learning a new programming language?
    \item What are the key steps to starting a small business?
    \item How can I improve my time management skills?
    \item What should I consider when choosing a graduate school program?
\end{enumerate}
\end{tcolorbox}

\begin{tcolorbox}[colback=orange!3, colframe=orange!40!black, title={\textbf{Category 3: Analysis / Comparison} (10 prompts)}, breakable, fonttitle=\small]
\small
\begin{enumerate}[leftmargin=*, itemsep=2pt]
    \item Compare renewable and non-renewable energy sources.
    \item What are the advantages and disadvantages of remote work?
    \item Compare Python and JavaScript as programming languages.
    \item What are the pros and cons of social media for society?
    \item Compare different types of database systems and their use cases.
    \item What are the benefits and risks of artificial intelligence?
    \item Compare democratic and authoritarian systems of government.
    \item What are the trade-offs between privacy and security in the digital age?
    \item Compare electric vehicles with traditional combustion engine cars.
    \item What are the advantages and disadvantages of online education?
\end{enumerate}
\end{tcolorbox}

\begin{tcolorbox}[colback=purple!3, colframe=purple!40!black, title={\textbf{Category 4: Technical / Detailed} (10 prompts)}, breakable, fonttitle=\small]
\small
\begin{enumerate}[leftmargin=*, itemsep=2pt]
    \item Explain how a neural network learns through backpropagation.
    \item Describe the process of DNA replication in cells.
    \item How does the TCP/IP protocol stack work?
    \item Explain the CAP theorem in distributed computing.
    \item How does a compiler translate source code into machine code?
    \item Describe how CRISPR gene editing technology works.
    \item Explain the principles behind public key cryptography.
    \item How does a recommendation system like Netflix's work?
    \item Describe how blockchain technology maintains a secure ledger.
    \item Explain how transformer models process natural language.
\end{enumerate}
\end{tcolorbox}

\section{Constraint Definitions}
\label{app:constraints}

\begin{tcolorbox}[colback=red!3, colframe=red!50!black, title={\textbf{Lexical Constraints}}, breakable, fonttitle=\small]
\small
Each constraint is appended verbatim to the user prompt as an additional sentence.

\medskip
\textbf{Punctuation-level constraints:}
\begin{itemize}[
    leftmargin=3cm,
    labelwidth=2.4cm,
    labelsep=0.5em,
    align=left,
    itemsep=4pt
]
    \item[\texttt{no\_comma}] ``Do not use any commas in your response.''
    \item[\texttt{no\_colon}] ``Do not use any colons in your response.''
    \item[\texttt{no\_semicolon}] ``Do not use any semicolons in your response.''
\end{itemize}

\medskip
\textbf{Pattern-level constraint:}
\begin{itemize}[
    leftmargin=3cm,
    labelwidth=2.4cm,
    labelsep=0.5em,
    align=left,
    itemsep=4pt
]
    \item[\texttt{no\_bullet}] ``Do not use bullet points, numbered lists, or dashes to create lists in your response. Write only in flowing prose paragraphs.''
\end{itemize}

\medskip
\textbf{Word-level constraints:}
\begin{itemize}[
    leftmargin=3cm,
    labelwidth=2.4cm,
    labelsep=0.5em,
    align=left,
    itemsep=4pt
]
    \item[\texttt{no\_the}] ``Do not use the word `the' in your response.''
    \item[\texttt{no\_discourse}] ``Do not use the words `however', `therefore', `moreover', `furthermore', or `additionally' in your response.''
\end{itemize}

\medskip
\textbf{Compositional constraints:}
\begin{itemize}[
    leftmargin=3cm,
    labelwidth=2.4cm,
    labelsep=0.5em,
    align=left,
    itemsep=4pt
]
    \item[\texttt{no\_comma\_colon}] ``Do not use any commas or colons in your response.''
    \item[\texttt{no\_comma\_bullet}] ``Do not use any commas in your response. Do not use bullet points, numbered lists, or dashes to create lists. Write only in flowing prose paragraphs.''
\end{itemize}
\end{tcolorbox}

\section{Judge Prompts}
\label{app:judge_prompts}

\subsection{Independent Scoring (Section~\ref{sec:eval})}

\begin{tcolorbox}[colback=gray!5, colframe=gray!60!black, title={\textbf{Independent Judge -- System Prompt}}, fonttitle=\small]
\small
You are an expert evaluator of AI assistant responses. Your job is to assess the QUALITY of a response to a user query, independent of any formatting constraints the user may have imposed.

Focus on the SUBSTANCE: Is the information accurate, comprehensive, and useful? Ignore formatting --- a response can be high quality even without bullet points or structured formatting.

You MUST output ONLY a valid JSON object with exactly four integer scores (1-10), nothing else.
\end{tcolorbox}

\begin{tcolorbox}[colback=gray!5, colframe=gray!60!black, title={\textbf{Independent Judge -- User Prompt Template}}, breakable, fonttitle=\small]
\small
Rate this AI assistant response on 4 dimensions (1-10 each).

\textbf{SCORING GUIDE:}
\begin{itemize}[leftmargin=1.5em, itemsep=1pt]
\item \textbf{1--3:} Poor. Missing key information, inaccurate, or unhelpfully vague.
\item \textbf{4--5:} Below average. Covers basics but lacks depth or misses important aspects.
\item \textbf{6--7:} Good. Covers the topic well with reasonable depth and accuracy.
\item \textbf{8--9:} Very good. Thorough, accurate, insightful, well-organized.
\item \textbf{10:} Exceptional. Could not meaningfully be improved.
\end{itemize}

\texttt{===== USER QUERY =====}\\
\texttt{\{prompt\}}

\texttt{===== AI RESPONSE =====}\\
\texttt{\{response\}}

\texttt{===== YOUR TASK =====}\\
Rate the response on:\\
1. informativeness: How much relevant, accurate information is provided?\\
2. depth: How detailed and thorough is the explanation?\\
3. clarity: How clear and understandable is the response?\\
4. helpfulness: Overall, how useful would this be to a typical user?

Output ONLY: \texttt{\{"informativeness": N, "depth": N, "clarity": N, "helpfulness": N\}}
\end{tcolorbox}

\subsection{Pairwise Comparison (Section~\ref{sec:main_results})}

\begin{tcolorbox}[colback=gray!5, colframe=gray!60!black, title={\textbf{Pairwise Judge -- System Prompt}}, fonttitle=\small]
\small
You are an expert evaluator comparing two AI assistant responses to the same question. Your job is to assess how comprehensive, detailed, and useful each response is RELATIVE to the other.

You MUST output ONLY a valid JSON object, nothing else.
\end{tcolorbox}

\begin{tcolorbox}[colback=gray!5, colframe=gray!60!black, title={\textbf{Pairwise Judge -- User Prompt Template}}, breakable, fonttitle=\small]
\small
Below are two responses (Response A and Response B) to the same user question. Rate EACH response on two dimensions (1-10).

\textbf{SCORING GUIDE} --- be discriminating, use the full scale:
\begin{itemize}[leftmargin=1.5em, itemsep=1pt]
\item \textbf{1--3:} Very brief or shallow. Misses most important aspects of the topic.
\item \textbf{4--5:} Covers basics but lacks detail, examples, or thorough explanation.
\item \textbf{6--7:} Good coverage with reasonable depth. Addresses key aspects.
\item \textbf{8--9:} Thorough and detailed. Provides examples, nuances, and structured explanation.
\item \textbf{10:} Exceptionally comprehensive. Could not meaningfully be improved.
\end{itemize}

\texttt{===== USER QUESTION =====}\\
\texttt{\{prompt\}}

\texttt{===== RESPONSE A =====}\\
\texttt{\{response\_a\}}

\texttt{===== RESPONSE B =====}\\
\texttt{\{response\_b\}}

\texttt{===== YOUR TASK =====}\\
Rate EACH response on:\\
1. comprehensiveness: How thoroughly does it cover the topic? Does it include examples, important details, edge cases, and structured explanation?\\
2. usefulness: How helpful would this be to someone trying to understand or act on this topic?

Output ONLY: \texttt{\{"response\_a\_comprehensiveness": N, "response\_a\_usefulness": N, "response\_b\_comprehensiveness": N, "response\_b\_usefulness": N\}}

\medskip
\emph{Note: The assignment of baseline/constrained responses to positions A and B is randomized for each pair to control for position bias.}
\end{tcolorbox}

\section{Two-Pass Qualitative Examples}
\label{app:twopass_examples}

We present a representative example from the two-pass experiment (\S\ref{sec:twopass}) on Qwen-2.5-7B-Instruct: the prompt ``Explain how vaccines work to protect against diseases'' under the constraint ``Do not use the word `the' in your response.''

\begin{tcolorbox}[colback=green!3, colframe=green!50!black, title={\textbf{Baseline Response} (unconstrained, 402 words)}, breakable, fonttitle=\small]
\small
Vaccines work by stimulating the immune system to recognize and fight specific pathogens (such as viruses or bacteria) without causing the disease itself. Here's a step-by-step explanation of how this process works:

1. \textbf{Introduction of Antigen:} The vaccine contains antigens, which are usually weakened or inactive forms of the pathogen (or parts of it). These antigens are introduced into the body, either through injection or another route.

2. \textbf{Immune System Recognition:} When the vaccine enters the body, the immune system recognizes these antigens as foreign substances. Specialized cells in the immune system, including dendritic cells, capture and process the antigens \ldots

\emph{[continues for 402 words with 7 numbered steps, bold headers, and detailed explanations]}
\end{tcolorbox}

\begin{tcolorbox}[colback=red!3, colframe=red!50!black, title={\textbf{Single-Pass Response} (constrained, 55 words, 86\% shorter)}, fonttitle=\small]
\small
Vaccines introduce small amounts of weakened or dead microorganisms into body, prompting immune system to recognize these invaders. Upon exposure, immune cells produce antibodies specific to antigens found on surface of these microorganisms. Afterward, if actual infection occurs, body quickly produces antibodies, neutralizing pathogens before illness develops. This process provides protection without causing disease.
\end{tcolorbox}

\begin{tcolorbox}[colback=blue!3, colframe=blue!50!black, title={\textbf{Two-Pass Response} (rewrite of baseline under constraint, 323 words, 80\% retention)}, breakable, fonttitle=\small]
\small
Vaccines operate by prompting immune system to identify and combat particular pathogens (like viruses or bacteria) without inducing disease. Here's a detailed breakdown of this process:

1. \textbf{Introduction of Antigen:} Vaccines contain antigens, typically weakened or inactive forms of pathogen (or parts of it). These antigens are introduced via injection or another method.

2. \textbf{Immune System Recognition:} When vaccine enters a body, immune system identifies antigens as foreign substances. Specialized cells, including dendritic cells, capture and process antigens \ldots

\emph{[continues for 323 words with same 7-step structure, preserving detail and organization]}
\end{tcolorbox}

The single-pass response collapses to a single paragraph of 55 words, losing all structure, examples, and detail, while the two-pass response preserves the original's numbered structure, bold headers, and substantive content while successfully avoiding the word ``the.''

\section{Constraint Satisfaction Rates}
\label{app:satisfaction}

Table~\ref{tab:satisfaction} reports constraint satisfaction rates for each model--constraint pair. Satisfaction is measured by automated checkers (character/word counters, regex pattern matching). Rates are generally high ($>$90\%), confirming that the quality collapse occurs among responses that successfully follow the constraint.

\begin{table}[h]
\caption{Constraint satisfaction rates (\%) across models. Measured on 120 responses per cell (40 prompts $\times$ 3 samples).}
\label{tab:satisfaction}
\centering
\small
\begin{tabular}{l cc}
\toprule
Constraint & Llama-3.1-8B-Inst. & Qwen-2.5-7B-Inst. \\
\midrule
No comma       & 99.2 & 97.5 \\
No colon       & 65.8 & 96.7 \\
No semicolon   & 100.0 & 83.3 \\
No bullet/lists& 100.0 & 100.0 \\
No ``the''     & 92.5 & 99.2 \\
No disc.\ markers & 98.3 & 91.7 \\
No comma+colon & 98.3 & 96.7 \\
No comma+bullet& 95.8 & 100.0 \\
\midrule
\textbf{Average} & \textbf{93.7} & \textbf{95.6} \\
\bottomrule
\end{tabular}
\end{table}

\paragraph{Note on colon satisfaction for Llama.} Llama-3.1-8B-Instruct achieves only 65.8\% satisfaction for the no-colon constraint because the model frequently generates headers with colons (e.g., ``Step 1: ...'') as part of its learned formatting templates. This is itself evidence of template dependence.

\section{Category Consistency}
\label{app:category}

The collapse holds across all four prompt categories (Table~\ref{tab:category}), ruling out domain-specific explanations. Technical prompts show the largest decline on Llama ($-$26.7\%), consistent with the expectation that structured, detail-rich content depends most on the formatting patterns disrupted by lexical constraints.

\begin{table}[h]
\caption{\textbf{Comprehensiveness change by prompt category} (GPT-4o-mini judge). The collapse is consistent across all four categories for all three instruct models, with no category showing positive change.}
\label{tab:category}
\centering
\small
\begin{tabular}{l ccc}
\toprule
Category & Llama Inst.\ $\Delta$\% & Qwen Inst.\ $\Delta$\% & Mistral Inst.\ $\Delta$\% \\
\midrule
Explanation  & $-$23.9 & $-$41.3 & $-$18.0 \\
How-to       & $-$22.2 & $-$41.5 & $-$7.7 \\
Analysis     & $-$20.5 & $-$36.6 & $-$15.7 \\
Technical    & $-$26.7 & $-$40.2 & $-$15.3 \\
\bottomrule
\end{tabular}
\end{table}

\section{GPT-4o-mini Per-Constraint Breakdown}
\label{app:gpt4omini}

\begin{table}[h]
\caption{\textbf{GPT-4o-mini (closed-weight) per-constraint results} (GPT-4o pairwise judge). 40 prompts per constraint. Comma bans are most damaging ($-$42\%), and average response length drops from 472 to 216 words ($-$54\%). The per-constraint pattern mirrors the open-weight models.}
\label{tab:gpt4omini}
\centering
\small
\begin{tabular}{l cccc}
\toprule
Constraint & Baseline $\bar{c}$ & Constr.\ $\bar{c}$ & $\Delta$\% & Win\% \\
\midrule
No comma       & 8.97 & 5.20 & $-$42.1 & 100 \\
No colon       & 8.97 & 6.25 & $-$30.4 & 100 \\
No semicolon   & 8.97 & 6.05 & $-$32.6 & 100 \\
No bullet/lists& 8.90 & 7.53 & $-$15.4 &  92 \\
No ``the''     & 8.97 & 6.10 & $-$32.0 & 100 \\
No disc.\ mkrs & 9.00 & 6.55 & $-$27.2 & 100 \\
\midrule
No comma+colon & 8.97 & 5.28 & $-$41.2 & 100 \\
No comma+bullet& 8.97 & 6.58 & $-$26.7 & 100 \\
\midrule
\textbf{Overall} & \textbf{8.97} & \textbf{6.19} & $\mathbf{-31.0}$ & $\mathbf{99.1}$ \\
\bottomrule
\end{tabular}
\end{table}

\section{MT-Bench Validation}
\label{app:mtbench}

To verify that our findings generalize beyond our evaluation prompts, we replicate the experiment on MT-Bench~\citep{zheng2024judging}, a standard 80-question benchmark spanning eight categories: writing, roleplay, reasoning, math, coding, extraction, STEM, and humanities. We test Llama-3.1-8B-Instruct and Qwen-2.5-7B-Instruct under three constraints (no comma, no ``the,'' no bullet), with GPT-4o as judge (240 pairwise comparisons per model).

\begin{figure}[h]
    \centering
    \includegraphics[width=\linewidth]{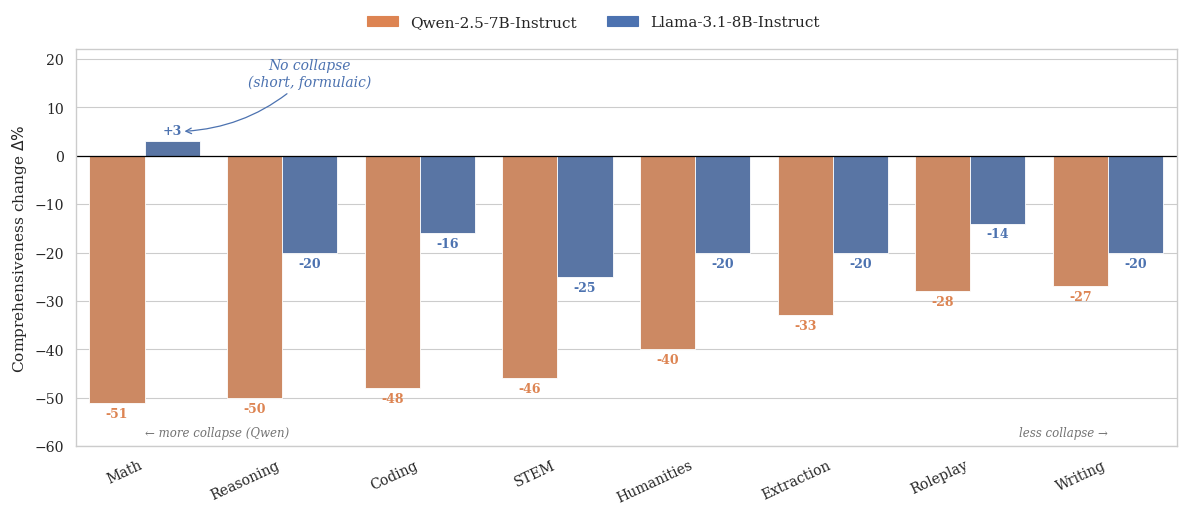}
    \caption{\textbf{Comprehensiveness change on MT-Bench by category} (GPT-4o pairwise judge). The collapse is consistent across all eight MT-Bench categories for both models. Llama math (+3\%) is the sole exception: its short, formulaic math responses do not rely on the formatting templates that collapse under constraints. Qwen collapses even on math ($-$51\%), consistent with its stronger template dependence.}
    \label{fig:mtbench}
\end{figure}

The results closely replicate our main findings (Figure~\ref{fig:mtbench}). Qwen shows $-$40.5\% overall comprehensiveness loss (89\% baseline win rate) on MT-Bench, nearly identical to the $-$39.9\% on our evaluation set. Llama shows $-$17.2\% (74\% win rate), consistent with its $-$23.4\% on our prompts. Per-constraint patterns are preserved: comma bans are most damaging (Llama $-$26\%, Qwen $-$65\%), followed by ``the'' (Llama $-$19\%, Qwen $-$42\%) and bullet bans (Llama $-$5\%, Qwen $-$12\%).

The category breakdown reveals that the collapse spans all task types, including those not represented in our evaluation set. STEM ($-$25\% to $-$46\%), reasoning ($-$20\% to $-$50\%), and coding ($-$16\% to $-$48\%) all show substantial degradation. The one exception is Llama on math (+3\%, chance-level win rate): its math responses are short and formulaic, not relying on the structured templates that collapse under constraints. Qwen, more aggressively template-dependent, collapses even on math ($-$51\%). This exception is itself informative: tasks whose optimal responses do not depend on structured formatting templates are naturally less vulnerable, confirming the template-dependence mechanism.

\section{Atomic Claim Coverage Analysis}
\label{app:coverage}

To address whether our pairwise judge responds to length and formatting rather than semantic coverage, we conduct a length-invariant content analysis. For 8 stratified prompts (2 per category) across all three open-weight instruct models (192 baseline--constrained pairs), we use GPT-4o to extract 11--20 atomic factual claims from each unconstrained response, then ask GPT-4o (with generous paraphrase matching) whether each baseline claim is conveyed in the corresponding constrained response. Coverage is defined as the fraction of baseline claims preserved; this metric is length-invariant by construction.

We define the \emph{length--coverage gap} as length retention minus atomic coverage. A gap near zero indicates length and content shed together, inconsistent with a verbosity-tax account.

\begin{table}[h]
\caption{\textbf{Atomic claim coverage analysis.} GPT-4o extracts factual claims from unconstrained responses and checks which survive in constrained responses (generous paraphrase matching). Coverage is length-invariant by construction. Overall gap $-0.8$pp is inconsistent with a pure verbosity account. 192 pairs, 3{,}355 atom checks.}
\label{tab:coverage_app}
\centering
\small
\begin{tabular}{l ccc}
\toprule
Model & Atomic Coverage & Length Retention & Gap \\
\midrule
Mistral-7B-Instruct    & 53.9\% & 59.2\% & $+$5.3 pp \\
Llama-3.1-8B-Instruct  & 57.4\% & 61.0\% & $+$3.6 pp \\
Qwen-2.5-7B-Instruct   & 38.2\% & 26.9\% & $-$11.3 pp \\
\midrule
\textbf{Overall}       & \textbf{49.8\%} & \textbf{49.0\%} & $\mathbf{-0.8}$ \textbf{pp} \\
\bottomrule
\end{tabular}
\end{table}

\paragraph{Results.} Constrained responses preserve only 49.8\% of baseline factual claims on average (Table~\ref{tab:coverage_app}). Three findings are inconsistent with a pure verbosity account. First, coverage and length retention move together at the aggregate level (overall gap $-0.8$pp). Second, Qwen exhibits a \emph{negative} gap ($-11.3$pp): severely shortened responses that are unusually dense per-word, yet still omit 62\% of baseline claims. Third, the \texttt{no\_bullet} constraint provides the cleanest anti-verbosity signal: averaged across models, responses retain 89\% of baseline length but only 58\% of baseline claims (gap $+31$pp).

\paragraph{Relation to pairwise severity.} Qwen is consistently the most fragile model across both pairwise comprehensiveness (\S\ref{sec:main_results}) and atomic coverage (38.2\% retention, versus 53.9--57.4\% for Mistral and Llama).

\paragraph{Caveats.} The extraction--matching pipeline is LLM-mediated end-to-end, so this analysis is supporting evidence alongside our pairwise, cross-judge, and human evaluations. The analysis is baseline-anchored and covers 8 stratified prompts per model. Per-constraint detail is provided in Table~\ref{tab:app_coverage_detail}.

\begin{table}[h]
\caption{Per-model, per-constraint atomic coverage and length retention. Each cell averages over 8 prompts. Gap = length retention $-$ coverage.}
\label{tab:app_coverage_detail}
\centering
\scriptsize
\begin{tabular}{l l ccc}
\toprule
Model & Constraint & Coverage & Length Ret. & Gap \\
\midrule
\multirow{8}{*}{Llama}
& No comma       & 55.0\% & 51.8\% & $-$3.2 pp \\
& No colon       & 54.6\% & 53.8\% & $-$0.8 pp \\
& No semicolon   & 56.9\% & 53.6\% & $-$3.3 pp \\
& No bullet/lists& 59.1\% & 92.5\% & $+$33.4 pp \\
& No ``the''     & 59.9\% & 49.2\% & $-$10.7 pp \\
& No disc.\ mkrs & 62.0\% & 64.3\% & $+$2.3 pp \\
& No comma+colon & 56.4\% & 51.2\% & $-$5.2 pp \\
& No comma+bullet& 55.2\% & 71.6\% & $+$16.4 pp \\
\midrule
\multirow{8}{*}{Qwen}
& No comma       & 33.9\% & 10.5\% & $-$23.4 pp \\
& No colon       & 33.2\% & 20.1\% & $-$13.1 pp \\
& No semicolon   & 42.4\% & 31.1\% & $-$11.3 pp \\
& No bullet/lists& 55.6\% & 69.0\% & $+$13.4 pp \\
& No ``the''     & 28.5\% & 15.0\% & $-$13.5 pp \\
& No disc.\ mkrs & 40.9\% & 29.4\% & $-$11.5 pp \\
& No comma+colon & 31.5\% & 11.7\% & $-$19.8 pp \\
& No comma+bullet& 39.7\% & 28.2\% & $-$11.5 pp \\
\midrule
\multirow{8}{*}{Mistral}
& No comma       & 49.9\% & 39.2\% & $-$10.7 pp \\
& No colon       & 52.3\% & 52.1\% & $-$0.2 pp \\
& No semicolon   & 56.7\% & 59.3\% & $+$2.6 pp \\
& No bullet/lists& 59.2\% & 105.5\% & $+$46.3 pp \\
& No ``the''     & 53.0\% & 52.1\% & $-$0.9 pp \\
& No disc.\ mkrs & 48.5\% & 61.9\% & $+$13.4 pp \\
& No comma+colon & 51.6\% & 32.3\% & $-$19.3 pp \\
& No comma+bullet& 59.8\% & 71.2\% & $+$11.4 pp \\
\midrule
\multicolumn{2}{l}{\textbf{Overall}} & \textbf{49.8\%} & \textbf{49.0\%} & $\mathbf{-0.8}$ \textbf{pp} \\
\bottomrule
\end{tabular}
\end{table}

\section{Coverage Analysis Judge Prompts}
\label{app:coverage_prompts}

\begin{tcolorbox}[colback=gray!5, colframe=gray!60!black, title={\textbf{Atom Extraction -- System Prompt}}, fonttitle=\small]
\small
You are an expert at extracting atomic factual claims from text. Your job is to decompose a response into its distinct informational atoms.

You MUST output ONLY a valid JSON object, nothing else.
\end{tcolorbox}

\begin{tcolorbox}[colback=gray!5, colframe=gray!60!black, title={\textbf{Atom Extraction -- User Prompt Template}}, breakable, fonttitle=\small]
\small
Below is a response to a user question. Extract the distinct factual claims or actionable points as a list of short, standalone sentences.

\textbf{RULES:}
\begin{itemize}[leftmargin=1.5em, itemsep=1pt]
\item Each claim should be ATOMIC: one fact, one idea per bullet.
\item Each claim should be STANDALONE: understandable without reading the others.
\item DO NOT include framing sentences, transitions, or meta-commentary (e.g., ``In conclusion\ldots'', ``First, we'll discuss\ldots'').
\item DO NOT include duplicated or paraphrased restatements of the same claim.
\item Aim for 8--15 atoms. If the response is shallow, extract fewer. If it is rich, extract more (up to 20).
\item Phrase each atom neutrally, in your own words -- not as a quote.
\end{itemize}

\texttt{===== USER QUESTION =====}\\
\texttt{\{prompt\}}

\texttt{===== RESPONSE =====}\\
\texttt{\{response\}}

\texttt{===== YOUR TASK =====}\\
Output ONLY: \texttt{\{"atoms": ["atom 1", "atom 2", ...]\}}
\end{tcolorbox}

\begin{tcolorbox}[colback=gray!5, colframe=gray!60!black, title={\textbf{Atom Matching -- System Prompt}}, fonttitle=\small]
\small
You are an expert at detecting whether a specific claim is present in a text, regardless of phrasing.

You MUST output ONLY a valid JSON object, nothing else.
\end{tcolorbox}

\begin{tcolorbox}[colback=gray!5, colframe=gray!60!black, title={\textbf{Atom Matching -- User Prompt Template}}, breakable, fonttitle=\small]
\small
Below is a response to a user question, followed by a specific claim. Your job: determine whether the response CONVEYS THIS CLAIM, even if phrased differently.

\textbf{MATCHING RULES:}
\begin{itemize}[leftmargin=1.5em, itemsep=1pt]
\item Answer YES if the response expresses the claim, even in different words, shorter form, or integrated into a larger sentence.
\item Answer YES even if the response merely implies the claim clearly.
\item Answer NO if the response omits, contradicts, or only partially covers the claim.
\item Be GENEROUS on paraphrase: same meaning = YES.
\item Be STRICT on substance: missing a key fact = NO.
\end{itemize}

\texttt{===== USER QUESTION =====}\\
\texttt{\{prompt\}}

\texttt{===== RESPONSE =====}\\
\texttt{\{response\}}

\texttt{===== CLAIM TO CHECK =====}\\
\texttt{\{claim\}}

\texttt{===== YOUR TASK =====}\\
Output ONLY: \texttt{\{"covered": "YES" or "NO", "reason": "brief justification (max 15 words)"\}}
\end{tcolorbox}

\section{Human Evaluation Protocol}
\label{app:human_eval}

\paragraph{Evaluator recruitment and ethics.} We recruit 10 evaluators with graduate-level education in STEM fields. Participation is anonymous and voluntary; all evaluators provide informed consent and are told that their ratings will be used to assess AI-generated response quality for a research study. No personally identifying information is collected. Evaluators are compensated for their time.

\paragraph{Task instructions.} Each evaluator receives a set of pairwise comparisons. For each comparison, the evaluator sees (1)~the original user question and (2)~two responses labeled Response A and Response B. One response is the unconstrained baseline and the other is the constrained response; the assignment to positions A and B is randomized for each pair. Evaluators are not informed which response is constrained, nor are they told the nature of the constraints applied. The evaluation is thus fully blinded with respect to experimental condition. Evaluators are instructed to judge each dimension independently.

\paragraph{Scoring criteria and anchor descriptions.} Each evaluator rates both responses on the following six criteria using a 1--10 scale, organized to separate information-level quality from surface-level presentation quality.

\begin{tcolorbox}[colback=gray!5, colframe=gray!50!black, title={\textbf{Information Criteria}}, fonttitle=\small, breakable]
\small

\textbf{1. Semantic Coverage} --- How many of the key ideas, subtopics, and important aspects of the question does the response address?
\begin{itemize}[leftmargin=1.5em, itemsep=1pt]
\item \textbf{1--3:} Covers only one or two aspects; misses most important subtopics.
\item \textbf{4--5:} Addresses some key points but omits several important aspects.
\item \textbf{6--7:} Covers most relevant subtopics with minor gaps.
\item \textbf{8--10:} Covers essentially all relevant subtopics and angles.
\end{itemize}

\medskip
\textbf{2. Comprehensiveness} --- Beyond mentioning topics, how much depth, detail, examples, and nuance does the response provide?
\begin{itemize}[leftmargin=1.5em, itemsep=1pt]
\item \textbf{1--3:} Superficial treatment; no examples or elaboration.
\item \textbf{4--5:} Some detail on a few points, but most topics covered only at surface level.
\item \textbf{6--7:} Reasonable depth on most points; some examples or elaboration.
\item \textbf{8--10:} Thorough treatment with examples, nuances, edge cases, and structured explanation.
\end{itemize}

\medskip
\textbf{3. Correctness} --- Is the information factually accurate?
\begin{itemize}[leftmargin=1.5em, itemsep=1pt]
\item \textbf{1--3:} Contains significant factual errors or fundamentally misleading claims.
\item \textbf{4--5:} Mostly correct but contains notable inaccuracies or oversimplifications.
\item \textbf{6--7:} Accurate with minor imprecisions that do not mislead.
\item \textbf{8--10:} Factually accurate throughout; claims are well-supported.
\end{itemize}

\medskip
\textbf{4. Helpfulness} --- Would this response actually help someone understand the topic or complete the task?
\begin{itemize}[leftmargin=1.5em, itemsep=1pt]
\item \textbf{1--3:} The reader would need to look elsewhere for useful information.
\item \textbf{4--5:} Provides a starting point but the reader would need supplementary sources.
\item \textbf{6--7:} Reasonably helpful; addresses the core question with actionable information.
\item \textbf{8--10:} The reader would feel well-informed and equipped to act after reading this.
\end{itemize}
\end{tcolorbox}

\begin{tcolorbox}[colback=gray!5, colframe=gray!50!black, title={\textbf{Surface Criteria}}, fonttitle=\small, breakable]
\small

\textbf{5. Verbosity (inverse)} --- How concise and efficient is the response? Higher scores indicate tighter writing with no unnecessary padding.

\emph{Important:} Brevity alone does not constitute conciseness. A short response that says little is empty, not concise. A long response packed with useful content is not verbose.

\begin{itemize}[leftmargin=1.5em, itemsep=1pt]
\item \textbf{1--3:} Excessively padded, repetitive, or filled with filler phrases.
\item \textbf{4--5:} Some unnecessary repetition or padding, but not egregious.
\item \textbf{6--7:} Reasonably concise; most content serves a purpose.
\item \textbf{8--10:} Every sentence contributes meaningfully; no wasted words.
\end{itemize}

\medskip
\textbf{6. Readability} --- How clear, well-organized, and easy to follow is the response?
\begin{itemize}[leftmargin=1.5em, itemsep=1pt]
\item \textbf{1--3:} Disorganized, hard to follow, or confusingly written.
\item \textbf{4--5:} Understandable but could be better organized or clearer.
\item \textbf{6--7:} Clear and well-organized; easy to follow.
\item \textbf{8--10:} Exceptionally clear; logical flow, effective use of structure.
\end{itemize}
\end{tcolorbox}

\paragraph{Methodological rationale.} The separation into information and surface criteria is the key design choice. If constrained responses score significantly lower on coverage, comprehensiveness, and helpfulness but comparably on verbosity and readability, the collapse reflects genuine information loss rather than a length-preference artifact.

\section{Per-Layer Probing Details}
\label{app:probing_layers}

Table~\ref{tab:probing_layers} provides the full per-layer Ridge regression $R^2$ for predicting response length from hidden states at the last prompt token, before generation begins.

\begin{table}[h]
\caption{\textbf{Linear probe results for instruction-tuned models.} Ridge regression $R^2$ for predicting response length from hidden states at the last prompt token. All three models peak at ${\sim}$50\% depth, with $R^2$ tracking collapse severity (Qwen $>$ Llama $>$ Mistral).}
\label{tab:probing_layers}
\centering
\small
\begin{tabular}{l cc cc cc}
\toprule
& \multicolumn{2}{c}{\textbf{Llama-3.1-8B-Inst.}} & \multicolumn{2}{c}{\textbf{Qwen-2.5-7B-Inst.}} & \multicolumn{2}{c}{\textbf{Mistral-7B-Inst.}} \\
\cmidrule(lr){2-3} \cmidrule(lr){4-5} \cmidrule(lr){6-7}
Depth & Idx & $R^2$ & Idx & $R^2$ & Idx & $R^2$ \\
\midrule
0\%    &  0 & 0.305 &  0 & 0.766 & 0 & $-$0.204 \\
25\%   &  8 & 0.735 &  7 & 0.883 & 8 & 0.429 \\
50\%   & 16 & \textbf{0.747} & 14 & \textbf{0.925} & 16 & \textbf{0.514} \\
75\%   & 24 & 0.730 & 21 & 0.921 & 24 & 0.508 \\
100\%  & 31 & 0.728 & 27 & 0.868 & 31 & 0.504 \\
\bottomrule
\end{tabular}
\end{table}

\section{Token-Level Strategy Divergence}
\label{app:divergence}

We run the model forward token-by-token on 5 prompts $\times$ 2 constraints, recording the top-50 token probability distribution at each of the first 20 generated positions for both the constrained and unconstrained prompts. We measure Jensen-Shannon divergence (JSD) and top-50 token overlap at each position.

\begin{table}[h]
\caption{\textbf{Token-level divergence} between constrained and unconstrained generation. JSD rises rapidly in the first 3--5 tokens and saturates across all three models. The model commits to a different response strategy within the opening tokens.}
\label{tab:divergence}
\centering
\small
\begin{tabular}{l cc cc cc}
\toprule
& \multicolumn{2}{c}{\textbf{Llama}} & \multicolumn{2}{c}{\textbf{Qwen}} & \multicolumn{2}{c}{\textbf{Mistral}} \\
\cmidrule(lr){2-3} \cmidrule(lr){4-5} \cmidrule(lr){6-7}
Position & JSD & Overlap & JSD & Overlap & JSD & Overlap \\
\midrule
1--3    & 0.544 & 35\% & 0.460 & 42\% & 0.512 & 36\% \\
4--10   & 0.983 &  8\% & 0.873 & 13\% & 0.780 & 22\% \\
11--20  & 0.995 &  9\% & 0.908 & 13\% & 0.896 & 15\% \\
\bottomrule
\end{tabular}
\end{table}

Qualitatively, unconstrained Llama opens with markdown formatting (e.g., ``\texttt{**Gradient Descent: A Simple Explanation**}'') while the constrained model opens with plain prose (``\texttt{Gradient descent is a way to find}''), committing to a fundamentally different response strategy in the very first token.

\section{Deployment Constraint Details}
\label{app:deployment_constraints}
\label{app:deployment_discussion}

\paragraph{Constraint text.} The four enterprise-grade constraints tested in \S\ref{sec:deployment}:
\begin{itemize}
    \item \textbf{Professional tone} (brand guidelines): ``Do not use exclamation marks, casual language, or informal expressions. Maintain a strictly professional and formal tone throughout your response.''
    \item \textbf{No preamble} (API efficiency / anti-sycophancy): ``Do not begin your response with a greeting, acknowledgment, or conversational opener such as `Certainly!', `Great question!', `I'd be happy to help!', `Sure!', or `Of course!'. Start directly with the first substantive sentence of your answer.''
    \item \textbf{Hedging language} (legal/compliance): ``Avoid making definitive or absolute claims. Use hedging language such as `may,' `might,' `could,' or `evidence suggests' instead of stating facts directly.''
    \item \textbf{Plain language} (accessibility): ``Write at a reading level accessible to a general audience. Avoid all technical jargon, acronyms, and complex sentence structures. Use simple, everyday words and short sentences.''
\end{itemize}

\paragraph{The professional tone tax.} Requesting a professional, formal tone is the default system prompt for virtually every corporate and customer-facing LLM deployment. This constraint places no restriction on length, vocabulary, structure, or factual depth, yet it costs the user 12--20\% of the response's comprehensiveness and shears off 17--34\% of word count. The fact that Qwen loses 148 words and 20\% of comprehensiveness simply because it was told not to be casual reveals how rigidly its factual knowledge is entangled with the conversational persona instilled by preference optimization.

\paragraph{The load-bearing preamble.} The no-preamble constraint restricts \emph{only} the opening tokens: the model is told not to begin with greetings or conversational openers such as ``Certainly!'' or ``Great question!'' No restriction is placed on the tone, vocabulary, structure, or content of the remainder of the response. Yet Qwen loses 40.4\% of comprehensiveness and 74.7\% of word count (448$\to$113 words), with the baseline preferred in 100\% of pairs. Llama loses 17.6\% (92\% baseline win rate). This connects directly to the token-level divergence analysis (Appendix~\ref{app:divergence}), which showed that models commit to a response strategy within the first 1--3 tokens. The RLHF-trained preamble is not filler; it functions as a conditional trigger that initializes the comprehensive response template. When the model cannot produce its trained opening, autoregressive decoding fails to route into the detailed, structured sub-policy, and the entire response collapses.

\paragraph{The hedging constraint.} Instructing a model to use ``may,'' ``might,'' or ``evidence suggests'' instead of definitive claims places no limit on the length, depth, or structural organization of the response. Yet this standard legal/compliance constraint triggers a 26\% collapse on \emph{both} models (Llama $-$26.4\%, Qwen $-$26.8\%), confirming that instruction-tuned models have coupled their factual knowledge to assertive, definitive formatting templates.

\paragraph{Plain language: a partially confounded upper bound.} The plain language constraint produces the largest degradation ($-$33.1\% Llama, $-$50.1\% Qwen), but this result requires careful interpretation. The constraint explicitly instructs the model to ``avoid all technical jargon'' and ``use simple, everyday words.'' For technical prompts, some comprehensiveness loss is an \emph{intended} consequence. The observed drop therefore represents an upper bound containing both unintended template collapse and intended complexity reduction. However, the severity of the length reduction (Qwen: 436 to 145 words, a 67\% drop) exceeds what jargon avoidance alone would predict.

\paragraph{Earlier 3-constraint pilot.} An earlier version of this analysis tested three deployment constraints (professional tone, hedging language, plain language) without the no-preamble constraint. The overall results ($-$24.1\% Llama, $-$32.4\% Qwen) closely matched the four-constraint results reported in the main paper, confirming robustness.

\section{Extended Discussion}
\label{app:extended_discussion}

This appendix preserves additional analysis and discussion from the main text that was relocated for space.

\paragraph{Underlying response changes.}
The comprehensiveness loss reflects dramatic structural changes. On Llama-3.1-8B-Instruct, banning commas reduces average response length by 57\% (from 685 to 297 tokens), unique content words by 41\%, and formatting richness (presence of bold text, headers, code blocks, bullet points) from 2.3 to 0.4 on a 0--4 scale. The model does not simply write the same content without commas; it produces a fundamentally different, minimal response.

\paragraph{Comprehensiveness and usefulness degrade in lockstep.}
Our pairwise judge collects two independent ratings: comprehensiveness and usefulness. Across all 13 model--judge configurations (10 instruct, 3 base), the per-pair Pearson correlation between the two is $r = 0.94$--$1.00$ (mean $0.99$), and their overall $\Delta\%$ values differ by at most 2.3 percentage points (mean 0.3pp). The blinded human evaluation (\S\ref{sec:human_eval}) further confirms this: human-rated helpfulness and comprehensiveness track within 0.3--3.2pp across all three models.

\paragraph{Why constraints sometimes help base models.}
Base model output is often unstructured and repetitive. Constraints like ``no bullet points'' or ``no commas'' can force the base model into more focused, coherent prose, inadvertently improving quality. The instruction-tuned model, whose unconstrained output is already its best-learned strategy, has no room for such accidental improvement.

\paragraph{What instruction tuning actually learns.}
Our results suggest that instruction tuning's apparent helpfulness is, at least in part, an artifact of learning a narrow distribution of response templates rather than developing a generalizable competence for providing thorough and accurate information. When any high-frequency token is banned, whether a formatting character (comma), a structural element (bullet points), or a common content word (``the''), the model's learned templates become inaccessible, and it defaults to a minimal response. The human evaluation (\S\ref{sec:human_eval}) provides the most direct evidence: blinded evaluators rate constrained responses as losing 16--44\% on information criteria while losing only 7--29\% on surface criteria. The collapse is not an evaluation artifact or a length-preference bias; it is a genuine loss of substantive content, confirmed by both automated and human judges. The atomic coverage analysis (\S\ref{sec:coverage}) reinforces this: only 49.8\% of baseline claims survive, and coverage tracks length retention rather than remaining high while length drops. The model does not know how to compress its answer; it substitutes a shorter, less informative response strategy instead. The base model probing control (\S\ref{sec:probing}) provides the representational counterpart: instruction tuning introduces not only the template-dependent strategy but also the predictive representational signature, which is entirely absent in base models.

\paragraph{From diagnostic probes to deployment constraints.}
A natural objection is that the specific lexical constraints we study are artificial: no real user asks ``explain gradient descent without commas.'' This is by design. Our lexical constraints serve as \emph{controlled diagnostic probes}, analogous to stress tests in structural engineering: no one drives a 100-ton truck across a pedestrian bridge in normal use, but if the bridge collapses under that load, it reveals something important about the bridge's structural integrity that normal foot traffic would never expose. Critically, we validate this analogy empirically (\S\ref{sec:deployment}): realistic deployment constraints produce collapse of comparable magnitude ($-$24\% to $-$32\%), with the hedging constraint alone causing 26\% degradation on both models despite placing no limit on response length, depth, or structure.

\paragraph{Implications for evaluation and deployment.}
The $6.7\times$ gap between independent and pairwise evaluation (\S\ref{sec:eval}) implies that any constrained generation system evaluated solely by independent scoring may carry undetected quality loss. The deployment constraint results make this concern concrete: merely requesting a professional tone, the default system prompt for most enterprise deployments, costs 12--20\% of comprehensiveness, and suppressing the conversational preamble costs up to 40\%. Combined with the GPT-4o-mini result (\S\ref{sec:scaling}), which shows a widely deployed commercial model losing 31\%, these findings suggest that the industry may be systematically overestimating the out-of-the-box utility of instruction-tuned models in constrained production environments. Practitioners should test their deployed models against the specific constraints they impose, using pairwise evaluation.

\paragraph{Toward robust instruction tuning.}
Our analysis suggests that constraint robustness should become an explicit training objective. The two-pass recovery result demonstrates feasibility: models possess the capability to produce comprehensive constrained output. The probing results localize the planning failure to middle-layer representations (${\sim}$50\% depth across all three architectures), identifying a consistent intervention site. The finding that $R^2$ tracks collapse severity (Figure~\ref{fig:probing_scatter}) suggests that reducing representational determinism at these layers may mitigate the collapse. The scaling results (\S\ref{sec:scaling}) indicate that simply training larger models is not a solution; if anything, models with higher baseline quality collapse more severely. Promising directions include constraint-augmented alignment training~\citep{yuan2024length}, representation-level interventions~\citep{stolfo2025improving}, and training with diverse surface-form constraints to decouple competence from any particular template.

\paragraph{Full limitations.}
Our diagnostic analysis is conducted on three 7--8B parameter open-weight models; we cannot determine whether the internal dynamics are identical in larger or closed-weight models, though the behavioral signature is consistent across all seven instruct models we evaluate (Table~\ref{tab:all_models}). Our primary evaluation uses LLM judges; the blinded human evaluation (\S\ref{sec:human_eval}) validates these findings, though the human study uses 10 evaluators, which, while sufficient given the large effect sizes (3--15$\times$ inter-rater variability), could be expanded in future work. The atomic coverage analysis (\S\ref{sec:coverage}) is LLM-mediated end-to-end and is baseline-anchored, measuring which baseline claims survive rather than whether constrained responses introduce new content; it is best read as supporting evidence alongside the human and LLM-judge evaluations. Our constraint set, while diverse, does not exhaustively cover all possible lexical constraints. The base model probing control uses 30 samples (vs.\ 120 for instruct), though the negative $R^2$ at all layers is unambiguous regardless of sample size. The probing analysis uses linear probes, which may underestimate the complexity of the underlying representation.

\section{Perplexity Analysis: Ruling Out OOD Likelihood Failure}
\label{app:perplexity}

To directly test whether constraint-induced collapse is driven by constrained text occupying a low-probability region of the language model's distribution, we compute the perplexity of baseline (unconstrained), single-pass (constrained, collapsed), and two-pass (constrained, comprehensive) responses under the corresponding \emph{base} (non-instruction-tuned) model. If comprehensive constrained text is fundamentally out-of-distribution (OOD), its perplexity under the base model should be dramatically elevated relative to unconstrained text. If the collapse is instead a planning failure specific to instruction tuning, the perplexity ratio should be modest.

Three behavioral observations also argue against the OOD hypothesis: (i)~two-pass recovery (\S\ref{sec:twopass}) shows that comprehensive constrained text is viable, (ii)~divergence analysis (Appendix~\ref{app:divergence}) shows coherent strategy switching rather than degenerate output, and (iii)~base models show no collapse under identical constraints (\S\ref{sec:instruct}). The perplexity analysis below provides direct mathematical confirmation.

\paragraph{Method.} For each of the 20 prompt--constraint pairs from the two-pass experiment (\S\ref{sec:twopass}), we compute the conditional perplexity $\text{PPL}(r \mid q) = \exp\!\bigl(-\frac{1}{N}\sum_{i=1}^{N} \log P_{\theta_\text{base}}(r_i \mid q, r_{<i})\bigr)$ for each response $r$ conditioned on the question $q$, using the base model parameters $\theta_\text{base}$. We evaluate on all three model families.

\begin{table}[h]
\caption{\textbf{Base-model perplexity of constrained text.} Conditional perplexity (given question) under the base model for unconstrained baseline, single-pass collapsed, and two-pass comprehensive responses. The two-pass/baseline ratio quantifies how OOD the comprehensive constrained text is. Llama and Mistral show ratios of $1.15$--$1.51\times$ (not OOD); Qwen shows a moderately elevated ratio ($2.54\times$), driven by the no-comma constraint, consistent with its partial two-pass recovery. 20 pairs per model.}
\label{tab:perplexity}
\centering
\small
\begin{tabular}{l ccc c}
\toprule
Model & Baseline PPL & Single-pass PPL & Two-pass PPL & TP/Baseline \\
\midrule
Llama-3.1-8B   & 1.8 & 2.5 & 2.0 & $1.15\times$ \\
Mistral-7B     & 1.9 & 2.2 & 2.8 & $1.51\times$ \\
Qwen-2.5-7B   & 2.0 & 5.6 & 5.1 & $2.54\times$ \\
\bottomrule
\end{tabular}
\end{table}

\paragraph{Results (Table~\ref{tab:perplexity}).} For Llama and Mistral, comprehensive constrained text (two-pass) has nearly identical perplexity to unconstrained text under the base model ($1.15\times$ and $1.51\times$ respectively). Comma-free or ``the''-free comprehensive prose is not fundamentally improbable; the base model assigns it comparable likelihood to standard text. This rules out OOD likelihood failure as the driver of single-pass collapse for these models and confirms the planning-failure interpretation.

Qwen shows a moderately elevated ratio ($2.54\times$), consistent with the partial two-pass recovery (59\%) reported in \S\ref{sec:twopass}. The elevation is driven primarily by the no-comma constraint ($3.28\times$), while no-``the'' is milder ($1.80\times$). This aligns with our behavioral finding that commas are deeply embedded in Qwen's generation patterns, introducing a secondary OOD component alongside the primary planning failure.

\paragraph{Single-pass perplexity is not lower.} A notable finding is that single-pass collapsed responses do \emph{not} have lower perplexity than two-pass comprehensive responses. For Qwen, single-pass perplexity ($5.6$) is actually \emph{higher} than two-pass ($5.1$). If the collapse were driven by autoregressive decoding seeking high-probability sequences, collapsed responses should have lower perplexity than comprehensive alternatives; they do not. The model's minimal-response strategy is a learned behavioral policy, not a likelihood-optimal decoding outcome. This is the strongest single piece of evidence against the OOD hypothesis: the instruction-tuned model actively routes into a sequence that the base model finds \emph{more} surprising than the comprehensive alternative it fails to produce.

\paragraph{Perplexity--recovery gradient.} The perplexity ratios map directly onto the behavioral recovery rates from the two-pass experiment (\S\ref{sec:twopass}):
\begin{center}
\small
\begin{tabular}{l cc}
\toprule
Model & TP/Baseline PPL & Two-Pass Recovery \\
\midrule
Llama-3.1-8B & $1.15\times$ & 96\% \\
Mistral-7B   & $1.51\times$ & 91\% \\
Qwen-2.5-7B  & $2.54\times$ & 59\% \\
\bottomrule
\end{tabular}
\end{center}
\noindent This gradient explains the variance in two-pass recovery that was previously an unexplained asymmetry across models. Qwen's partial recovery is a direct consequence of its instruction-tuning recipe pushing comma-free comprehensive prose further from the base distribution than Llama or Mistral. The collapse is primarily a planning failure for all three models, with a secondary OOD component whose magnitude tracks the specific instruction-tuning recipe's template dependence.

\end{document}